\documentclass{article}

\PassOptionsToPackage{numbers,compress}{natbib}
 \usepackage[preprint]{neurips_2026}

\usepackage[utf8]{inputenc} % allow utf-8 input
\usepackage[T1]{fontenc}    % use 8-bit T1 fonts
\usepackage{hyperref}       % hyperlinks
\usepackage{url}            % simple URL typesetting
\usepackage{booktabs}       % professional-quality tables
\usepackage{amsfonts}       % blackboard math symbols
\usepackage{nicefrac}       % compact symbols for 1/2, etc.
\usepackage{microtype}      % microtypography
\usepackage{xcolor}         % colors
\usepackage{graphicx}
\usepackage{array}
\usepackage{colortbl}
\usepackage{tabularx}
\usepackage{makecell}
\usepackage{amsmath}
\usepackage{multirow}
\usepackage{enumitem}
\usepackage{subcaption}
\usepackage{tikz}
\usepackage{placeins}
\usetikzlibrary{shapes.geometric, arrows.meta, positioning, fit, backgrounds}
\usepackage{overpic}
\usepackage{pgfplots}
\pgfplotsset{compat=1.18}
\usepackage{geometry}
\usepackage{pifont}
\newcommand{\cmark}{\ding{51}}
\newcommand{\xmark}{\ding{55}}
\hypersetup{
  colorlinks = true,
  linkcolor  = blue!70!black,
  citecolor  = green!50!black,
  urlcolor   = blue!70!black,
}

\usepackage[skins]{tcolorbox}
\definecolor{promptbg}{RGB}{227,242,253}
\definecolor{promptborder}{RGB}{30,136,229}
\newtcolorbox{promptenv}[1]{%
  colback=promptbg,
  colframe=promptborder,
  colbacktitle=promptborder,
  coltitle=white,
  fonttitle=\bfseries\small,
  title={#1},
  rounded corners,
  arc=4pt,
  boxrule=1.2pt,
  left=6pt, right=6pt, top=4pt, bottom=4pt,
  before upper={\small},
}
\newcommand{\promptbox}[2]{%
  \par\medskip\noindent
  \begin{promptenv}{#1}#2\end{promptenv}
  \par\medskip
}

\newcommand{\tni}{\textsc{TNI}}

\newcommand{\gcmark}{\textcolor{green!55!black}{\ding{51}}}
\newcommand{\rxmark}{\textcolor{red!75!black}{\ding{55}}}
\newcommand{\partialmark}{\textcolor{orange!85!black}{$\sim$}}

\newcommand{\logoa}{\raisebox{-0.18\height}{\includegraphics[height=8pt]{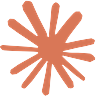}}}
\newcommand{\logog}{\raisebox{-0.18\height}{\includegraphics[height=8pt]{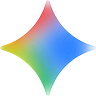}}}
\newcommand{\logoo}{\raisebox{-0.18\height}{\includegraphics[height=8pt]{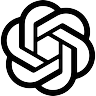}}}
\newcommand{\logoq}{\raisebox{-0.18\height}{\includegraphics[height=8pt]{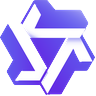}}}

\newcolumntype{C}[1]{>{\centering\arraybackslash}p{#1}}
\newcolumntype{L}[1]{>{\raggedright\arraybackslash}p{#1}}

\definecolor{plannerblue}{RGB}{52, 120, 200}
\definecolor{generatorgreen}{RGB}{40, 160, 80}
\definecolor{evaluatorred}{RGB}{200, 60, 60}
\definecolor{lightgray}{RGB}{240, 240, 240}
\definecolor{darkgray}{RGB}{80, 80, 80}
\definecolor{checkgreen}{RGB}{0, 150, 60}
\definecolor{crossred}{RGB}{180, 30, 30}
\usepackage[dvipsnames]{xcolor}

\title{\raisebox{-0.35\height}{\includegraphics[height=0.95cm]{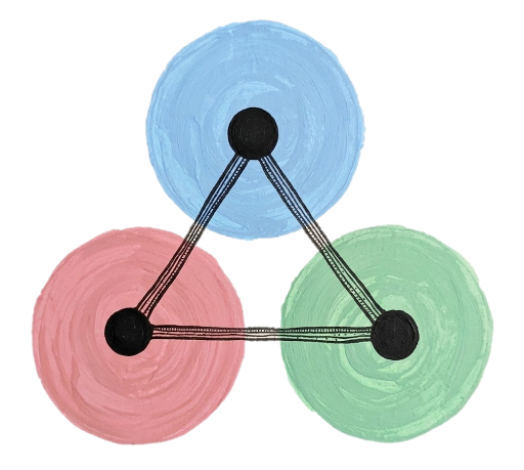}}\hspace{0.1cm}TeamBench: Evaluating Agent Coordination under Enforced Role Separation}

\author{
Yubin Kim$^{1,2}$ \quad
Chanwoo Park$^{1}$ \quad
Taehan Kim$^{4}$ \quad
Eugene Park$^{1}$ \quad
Samuel Schmidgall$^{3}$ \\
Salman Rahman$^{2}$ \quad
Chunjong Park$^{3}$ \quad
Cynthia Breazeal$^{1}$ \quad
Xin Liu$^{2}$ \quad
Hamid Palangi$^{2}$ \\
Hae Won Park$^{1}$ \quad
Daniel McDuff$^{2}$ \\ \\
$^{1}$MIT \quad
$^{2}$Google Research \quad
$^{3}$Google DeepMind \quad
$^{4}$Independent Researcher \\[6pt]
{\footnotesize \includegraphics[height=9pt]{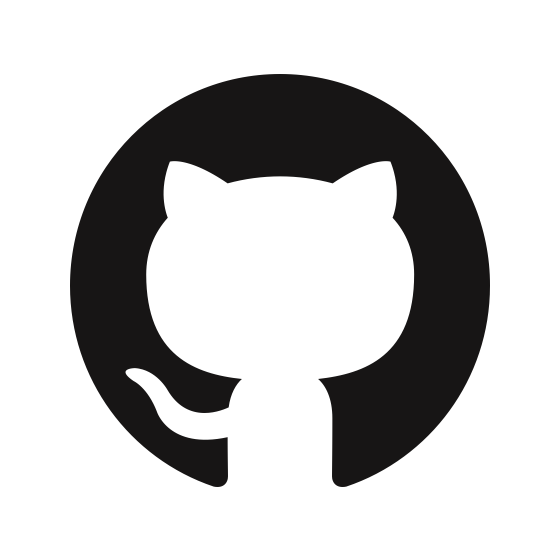}~~
\href{https://github.com/ybkim95/TeamBench}{GitHub}} \hspace{6pt}
{\footnotesize \includegraphics[height=9pt]{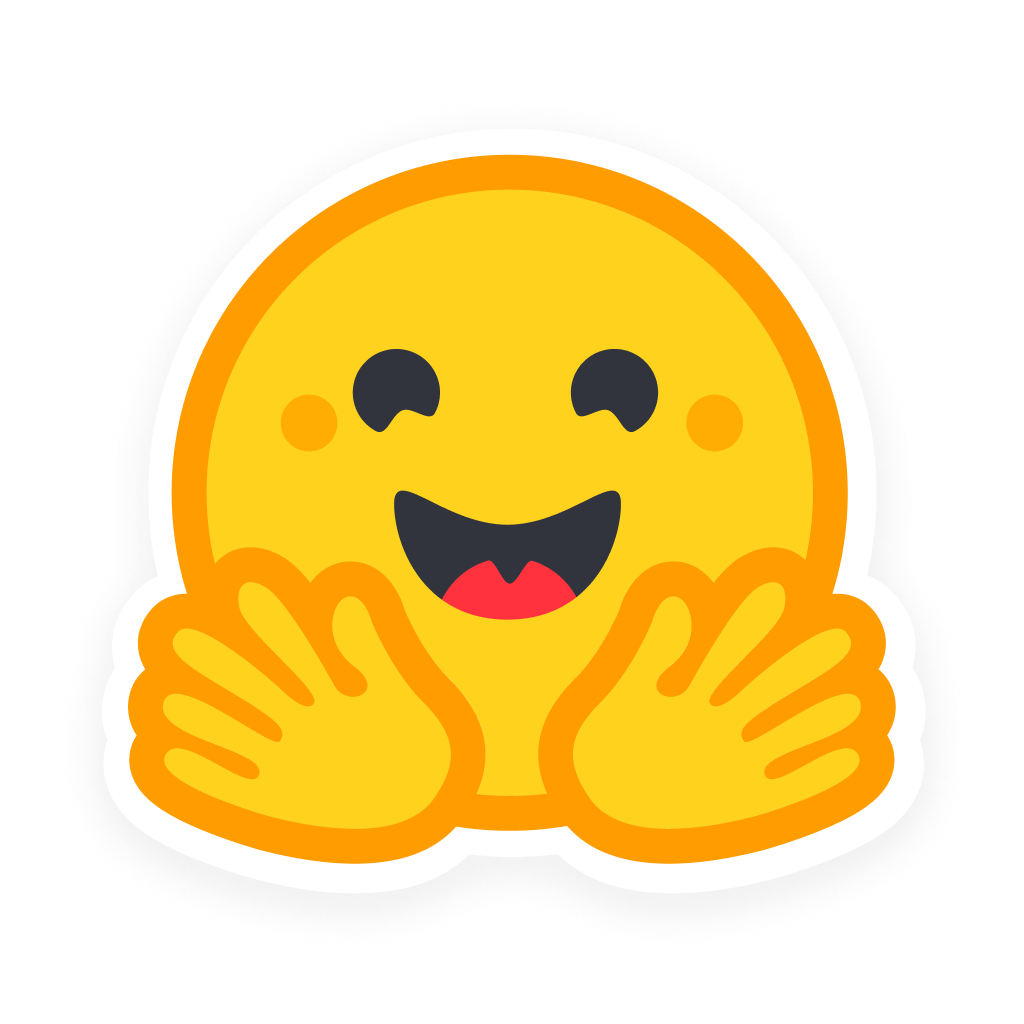}~~
\href{https://huggingface.co/datasets/ybkim95/teambench}{Dataset}} \hspace{6pt}
{\footnotesize \includegraphics[height=9pt]{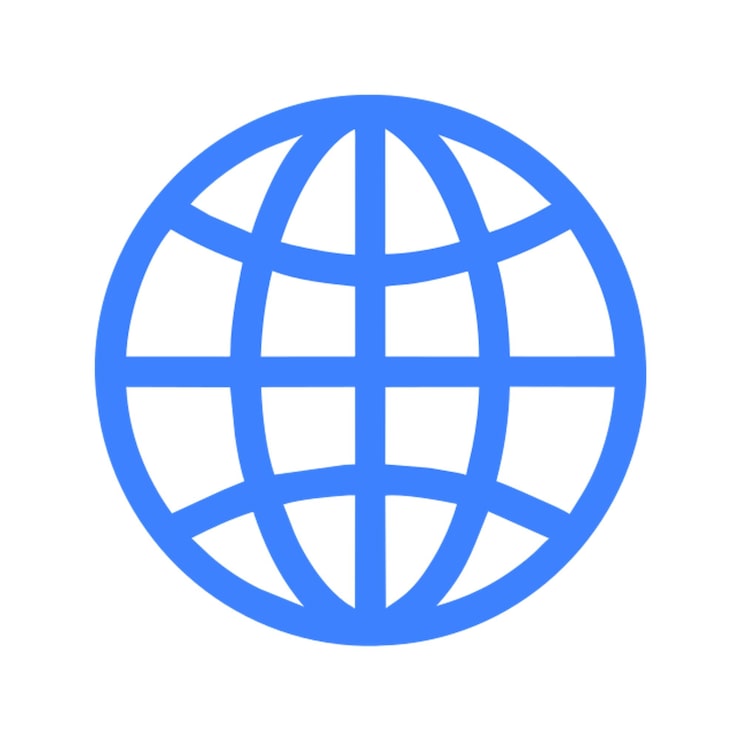}~~
\href{https://teambench.github.io}{Website}}
}

\begin{document}

\maketitle

\begin{abstract}
Agent systems often decompose a task across multiple roles, but these roles are typically specified by prompts rather than enforced by access controls. Without enforcement, a team pass rate can mask whether agents actually coordinated or whether one role effectively did another role's work. We present \textsc{TeamBench}, a benchmark with 851 task templates and 931 seeded instances for evaluating agent coordination under operating system-enforced role separation. \textsc{TeamBench} separates specification access, workspace editing, and final certification across Planner, Executor, and Verifier roles, so that no role can read the full requirements, modify the workspace, and certify the final answer. Prompt-only and sandbox-enforced teams reach statistically indistinguishable pass rates, but prompt-only runs produce 3.6 times more cases where the verifier attempts to edit the executor's code. Verifiers approve 49\% of submissions that fail the deterministic grader, and removing the verifier improves mean partial score in the ablation. Team value is also conditional. Teams benefit when single agents struggle, but hurt when single agents already perform well. A 40-session human study under the same role separation shows that our benchmark exposes interaction patterns that pass rate misses. Solo participants work through the task directly, human participants paired with agents often collapse into quick approval, and human teams spend more effort coordinating missing information across roles. Our dataset, code and implementation details can be found at \url{https://teambench.github.io}.
% \textsc{TeamBench} turns role separation from a prompt-level assumption into an evaluable property of agent systems.
\end{abstract}

%% ============================================================
%% 1. INTRODUCTION
%% ============================================================
\section{Introduction}
\label{sec:intro}

\begin{figure}[t!]
    \centering
    {\includegraphics[width=\linewidth]{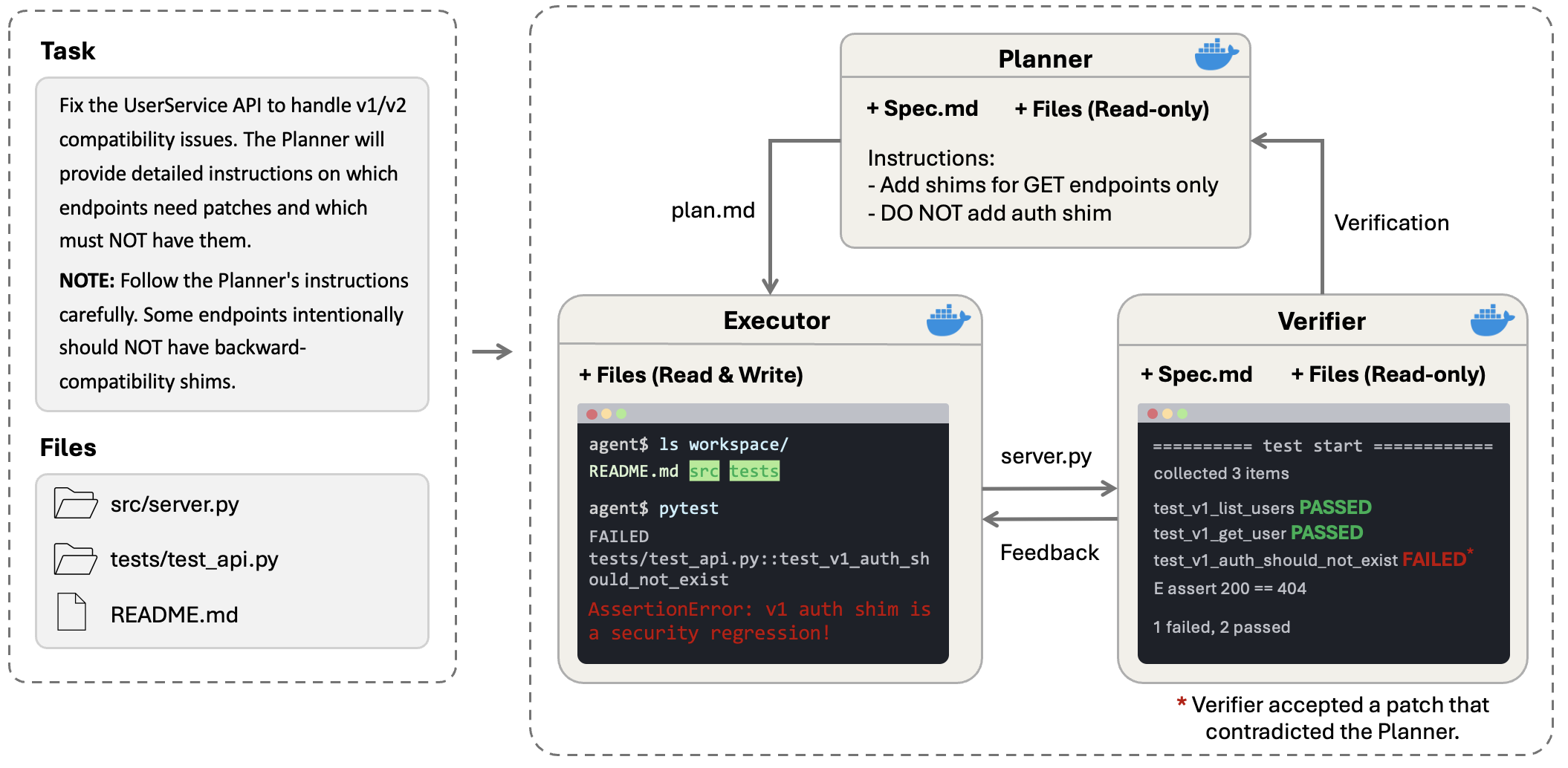}}
    \caption{\textsc{TeamBench} evaluates agents under enforced role separation. The Planner reads the full requirements, the Executor edits the workspace, and the Verifier issues the final verdict. Sandboxes restrict which files each role can read or write. The example shows a role failure, where the Verifier accepts an implementation that violates a Planner constraint.}
    \label{fig:teaser}
\end{figure}

Agent-based systems built on large language models (LLMs) often split tasks across different roles and measure gains over a single agent~\citep{hong2024metagpt, wu2023autogen, qian2024chatdev, li2023camel, zhuge2024gptswarm, kim2025towards}. Yet it is often unclear whether those gains come from coordination among roles or from one model effectively carrying several roles within the same run. A team that adds a verification step needs to know whether the Verifier is an independent quality gate or simply another editing pass. In many benchmarks, different roles are assigned through prompts to the same backbone model, and the harness does not prevent that model from planning, editing, and certifying the same solution. Recent works raise related concerns about role collapse, benchmark validity, and multi-agent failure modes~\citep{anthropic2025multiagent, cemri2025whyfail, zhu2025abc, li2025hiddenbench}. SWE-Bench~\citep{jimenez2024swebench} and Terminal-Bench~\citep{terminalbench2026} ask whether an agent can solve realistic tasks. \textsc{TeamBench} asks when a Planner, Executor, and Verifier decomposition produces useful coordination, and when it only adds overhead without genuine coordination.

We study this question by enforcing role separation with operating system permissions. Each role runs in a separate container with access only to the files and tools needed for that role. The Planner reads the full requirements but cannot modify the workspace. The Executor can edit and test the workspace, but only receives a summarized brief rather than the full specification. The Verifier reviews the requirements and the Executor’s evidence before approving the final submission. No role can simultaneously read the full requirements, modify the workspace, and certify the final answer. This design makes coordination necessary because information must move across roles for the team to complete the task.

\textsc{TeamBench} contains $851$ task templates that expand to $931$ seeded evaluation instances. The tasks span 19 base categories, with 21 refined categories for the leaderboard. Each task has a generator that produces a deterministic workspace from a seed, allowing instances to refresh without retiring the benchmark. We evaluate models on a stratified 90-task subset and also report \textsc{TeamBench-Verified}. Beyond the leaderboard, the evaluation suite includes Planner and Verifier ablations, a 27-configuration cross-provider role-mixing ablation, a prompt-only versus enforced-role comparison, and a human study with the same role boundaries and deterministic graders.

Several findings across the experiments include: (i) In the role-mixing ablation, Verifiers approve 49\% of submissions that fail the deterministic grader, and removing the Verifier improves partial score in the reference ablation. (ii) Teams help most in the lowest Solo-score quintile, but hurt on easier tasks where Solo already performs well. (iii) Prompt-only and sandbox-enforced roles reach statistically indistinguishable pass rates, while prompt-only runs produce 3.6 times more cases where the Verifier rewrites the Executor's code. (iv) In the human study, Solo participants work through the task directly, Hybrid (participant paired with agents) sessions often collapse into quick approval, and human teams spend more effort coordinating missing information across roles. These results indicate that the same pass rate can hide different role and verification behavior, and coordination costs.

Our primary contributions are:
\begin{enumerate}[leftmargin=1em,itemsep=1pt,topsep=0pt]
    \item \textbf{Role-separated coordination benchmark.}
    We introduce \textsc{TeamBench}, which uses operating-system permissions to separate
    specification access, workspace editing, and final certification.

    \item \textbf{Matched ablations for role value.}
    We run matched ablation under Solo, Restricted, Full Team, No-Plan, and No-Evaluate
    conditions, isolating when planning, verification, and missing specification access
    improve performance.

    \item \textbf{Verifier failure as a bottleneck.}
    Verifiers approve 49\% of grader-failing submissions, and removing the Verifier
    improves mean partial score in the reference ablation.

    \item \textbf{Coordination behavior beyond pass rate.}
    Team value depends on Solo capability, prompt-only roles hide more Verifier code-edit attempts,
    and our human study suggests distinct Solo, Hybrid, and Team interaction patterns
    under the same role boundaries.
\end{enumerate}

%% ============================================================
%% 2. TEAMBENCH
%% ============================================================
\section{TeamBench}
\label{sec:teambench}

\textsc{TeamBench} isolates three roles that prompt-only evaluations blend. It measures the marginal contribution of each role, the effect of assigning different providers to different roles, and how team value changes with task difficulty. To measure these effects, we enforce the role boundaries by the harness rather than via the prompt. Table~\ref{tab:benchmark-comparison} compares \textsc{TeamBench} with existing agent benchmarks. 

\subsection{Role Separation}
\label{sec:roles}

Each role runs in a separate container with only the files and tools it needs. The Planner receives the full requirements but cannot edit or execute code. The Executor modifies the solution workspace and runs commands but never sees the full requirements, only a brief. The Verifier reads the full requirements and read-only evidence from the Executor, then issues the final pass-or-fail verdict. Because no role has all three permissions, a successful run requires information to move across roles. The three roles separate task understanding, implementation, and validation. Removing either the Verifier or the Planner gives an ablation condition (Table~\ref{tab:conditions}). Implementation details, including container boundaries, file paths, and attestation format are in Appendix~\ref{app:role-permissions}.

\subsection{Task Construction}
\label{sec:tasks}

\begin{figure}[t]
    \centering
    \includegraphics[width=\linewidth]{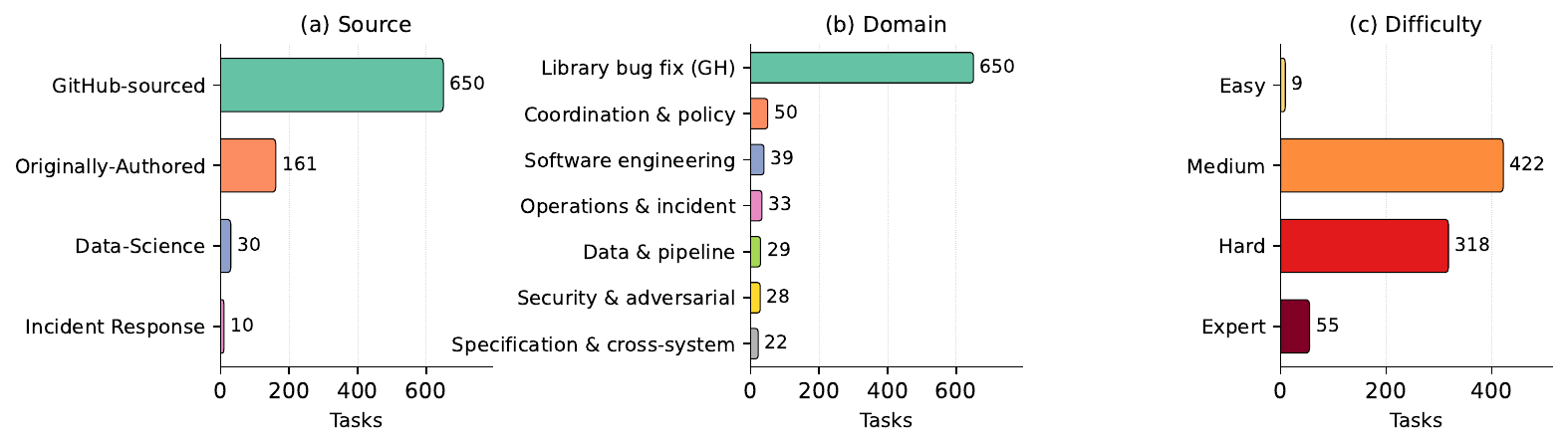}
    \caption{\textbf{\textsc{TeamBench} composition.} \textbf{(a)} Source distribution by origin class. \textbf{(b)} Domain distribution across the 851 task templates. \textbf{(c)} Difficulty distribution over the 804 templates with grader-check counts. The remaining 47 templates are Unrated due to the grader-check count was not available.}
    \label{fig:full-pool-pie}
\end{figure}

A coordination benchmark needs tasks where the specification contains information that the workspace alone does not reveal. If the workspace alone contains enough information for an Executor-only agent to solve the task, the task does not require coordination. We curate $851$ templates with this property. Each template includes a deterministic script grader and expands to one or more seeded evaluation instances, for 931 instances total. We curate 161 tasks with critical constraints available only in the full requirements. We adapt 650 GitHub bug reports from active open source repositories, including Flask, Django, NumPy, SciPy, and Keras. We also include 30 data science tasks built on canonical public datasets~\citep{dua2019uci} and 10 incident response tasks adapted from public post mortems. The pool spans $19$ base categories covering security patching, data-pipeline repair, distributed systems debugging, cryptographic correctness, adversarial specification traps, and more. The benchmark includes many challenging tasks because \S\ref{sec:capability} tests whether teams help most when single agents struggle. Figure~\ref{fig:full-pool-pie} shows the composition.

Tasks span five coordination failure modes. Relay tasks require the Planner to pass specification details that the Executor cannot infer from the workspace. Adversarial-trap tasks place plausible but incorrect content in the workspace. Open-ended tasks allow multiple valid implementations, but deterministic checks define which outputs pass. Discovery tasks hide data-quality or API issues that the Executor must surface through active exploration. Synthesis tasks require correlating evidence across multiple documents. In the 161 authored templates, critical constraints appear only in the full specification. They are absent from the brief and workspace, so the Executor needs information from the Planner. Per-origin counts and file layout are in Appendix~\ref{app:task-construction}.

Every task deploys with a generator that produces deterministic but distinct workspace files from different random seeds. Generators vary task-relevant parameters (configuration values, API field names, bug locations) while preserving structural complexity, which protects against value memorization. 

% Held-out seeds are reserved for the leaderboard and are not present in the public release.

\subsection{Ablation Conditions}
\label{sec:tni}

We run the same task under five conditions (Table~\ref{tab:conditions}) that isolate the marginal contribution of each role. Solo is a single agent with the full spec, workspace, and four tools. Restricted is the same single agent without access to the full specification.  The three team conditions (Full Team, Team, No Plan and Team, No Evaluate) then add or remove the Planner and Verifier.

\begin{table}[t]
\centering
\caption{Ablation conditions, with which agent holds each capability. \textit{Spec} = read full specification, \textit{Edit} = modify workspace and execute commands, \textit{Attest} = write the closing attestation. $^\dagger$In Team-No-Plan, the Executor sees only the brief, and the Verifier holds the full spec for compliance checking.}
\label{tab:conditions}
\small
\begin{tabular}{llccc}
\toprule
Condition & Agents & Spec & Edit & Attest \\
\midrule
Solo               & one agent (full access)              & \checkmark & \checkmark & \checkmark \\
Restricted         & one agent (Executor-only tools)      & --         & \checkmark & --         \\
Full Team          & Planner + Executor + Verifier        & Planner    & Executor   & Verifier   \\
Team, No Plan      & Executor + Verifier                  & Verifier$^\dagger$ & Executor & Verifier   \\
Team, No Evaluate  & Planner + Executor                   & Planner    & Executor   & --         \\
\bottomrule
\end{tabular}
\end{table}

We define the \textbf{Teamwork Necessity Index (TNI)} as the fraction of the Solo versus Restricted gap recovered by the team. Intuitively, TNI asks how much performance is recovered when the missing requirements must be relayed through the team:
\begin{equation}
\tni = \frac{S_{\text{team}} - S_{\text{restricted}}}{\max(\epsilon,\, S_{\text{solo}} - S_{\text{restricted}})},
\label{eq:tni}
\end{equation}
with $\epsilon = 0.05$ to avoid instability when the Solo--Restricted gap is near zero.
$\tni = 1$ indicates that the team fully recovers the single-agent advantage, while
$\tni > 1$ exceeds it. We summarize TNI only over tasks where
$S_{\text{solo}} - S_{\text{restricted}} > \epsilon$, since smaller Solo--Restricted
gaps do not provide a meaningful test of teamwork necessity. We additionally report
Planning Value $= S_{\text{full}} - S_{\text{no\_plan}}$ and Verification Value
$= S_{\text{full}} - S_{\text{no\_verify}}$, and classify tasks as HIGH-TNI,
TEAM-HELPS, NEUTRAL, or TEAM-HURTS using a $\pm 0.05$ band.

\subsection{Role Mixing}
\label{sec:role-mixing}

If Planner, Executor, and Verifier roles require distinct capabilities, then the best model for one role need not be the best model for another. We therefore treat model assignment as an experimental variable rather than fixing one model across all roles, as much prior agent-based work does. The protocol enumerates every assignment of three models to three roles, giving $27$ configurations, restricted to three LLM families (Anthropic, Google, OpenAI) with one model per family to keep the grid tractable. Each configuration runs all $25$ tasks of a stratified subset. We use a compact code for configurations, with one letter for the provider assigned to each role. For instance, \texttt{PGEAVO} runs the Google Planner, the Anthropic Executor, and the OpenAI Verifier. 

%% ============================================================
%% 3. EXPERIMENTS AND RESULTS
%% ============================================================
\section{Experiments and Results}
\label{sec:experiment}

The experiments address three questions. 1) What is the marginal contribution of each role under structural enforcement? 2) Does pass rate depend on which provider fills which role? 3) In which task regimes does the team provide positive uplift, and where does it hurt? Appendix~\ref{app:study-counts} lists the studies and the count behind each one.

\subsection{Setup}
\label{sec:exp-studies}

The leaderboard evaluates $13$ models across four families (Anthropic, Google, OpenAI, and Alibaba). The cross provider grid uses one compact frontier model from each commercial family. Every role is a tool-calling loop in its own sandbox with tools. Each task includes a deterministic script grader. A binary pass requires every check to pass, following SWE-Bench~\citep{jimenez2024swebench} and Terminal-Bench~\citep{terminalbench2026}. Held-out seeds are reserved for leaderboard refresh. Comparisons use paired bootstrap ($10{,}000$ iterations) with Wilson $95\%$ CIs. The enforcement study uses McNemar with Holm-Bonferroni. Reproducibility details are listed in Appendix~\ref{app:reproducibility}.

\subsection{Main Results}
\label{sec:lb90}

\begin{figure}[t]
    \centering
    \includegraphics[width=\linewidth]{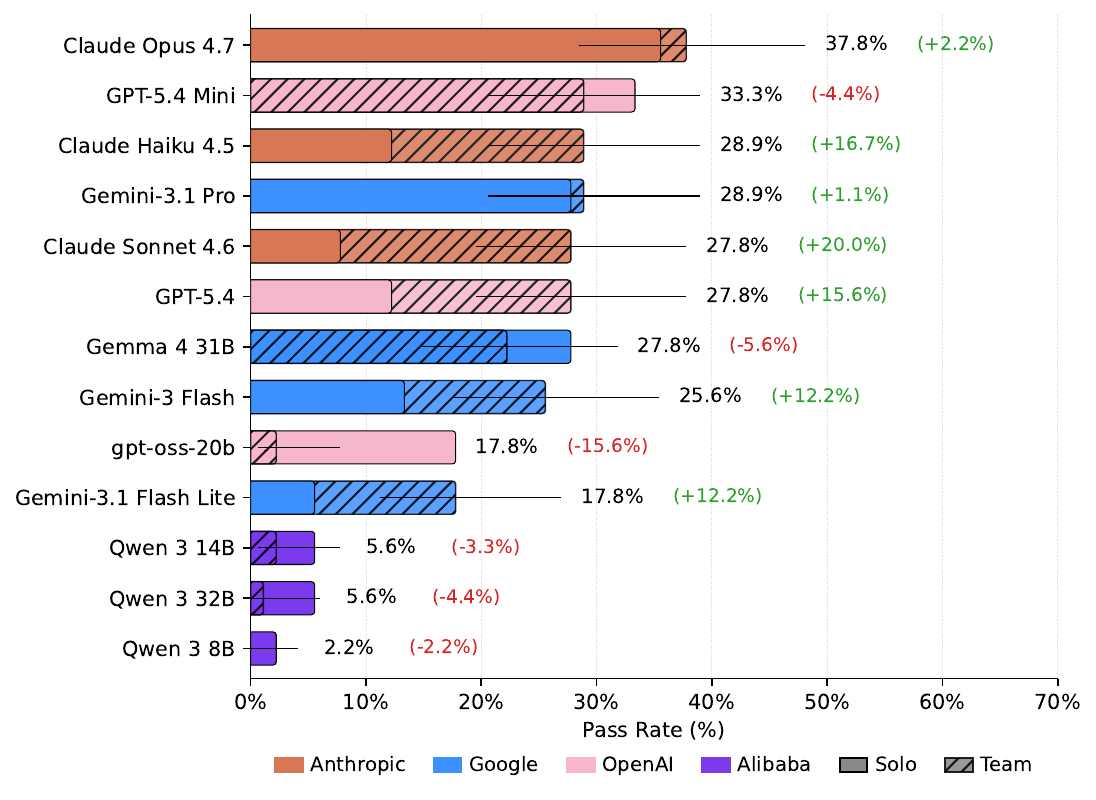}
    \caption{\textbf{\textsc{TeamBench} leaderboard.} Rows are sorted by $\max($Solo, Team$)$ descending so the row order tracks each model's best demonstrated capability. The bar shows both Solo (solid family color) and Team (hatched), and the right-side label gives that maximum percentage with the parenthesized delta to the other condition in green (\textcolor{ForestGreen}{$+$}, team helps) or red (\textcolor{red}{$-$}, team hurts). The shorter of the two bars is drawn on top so both endpoints are visible. Whiskers are Wilson $95\%$ CIs on the Team rate.}
    \label{fig:leaderboard}
\end{figure}

Figure~\ref{fig:leaderboard} shows that \textsc{TeamBench} remains difficult even for frontier models. Claude Opus 4.7 is the strongest model in both settings,
with $37.8\%$ on Full Team and $35.6\%$ on Solo. Across models, however, the
gap between Full Team and Solo changes direction with Solo performance. Models
with low Solo scores often improve under the team setting, including Sonnet 4.6
($+20.0$), Haiku 4.5 ($+16.7$), GPT-5.4 ($+15.6$), and Gemini-3 Flash
($+12.2$). Models with stronger Solo performance show little gain or lose accuracy,
including Opus 4.7 ($+2.2$), GPT-5.4 Mini ($-4.4$), and Gemma 4 31B ($-5.6$).
This pattern suggests that role separation helps when a single agent lacks enough
planning context to make progress, but can add failure opportunities once the Solo
agent already solves much of the task.

We follow the counting convention in Table~\ref{tab:lb90-leaderboard}. A run that passes all structural checks but fails only the final attestation check is counted as a pass,
because the attestation file records metadata compliance rather than task quality.
The Qwen 3 family and gpt-oss-20b score below $18\%$ on Solo and fall further
in the team setting, mainly because of tool-call errors and context-window overflow
rather than task reasoning failures (Appendix~\ref{app:open-source}).

\subsection{The Contribution of Each Role}
\label{sec:ref-results}

\textbf{Verifiers reduce average score in the reference ablation.} In the 155-task reference ablation with Gemini-3 Flash, which fixes the model across all five conditions, removing the Verifier from the Full Team raises mean partial score by $5.5$ points, and the per-task verification value averages
$-5.8$ points. This decrease could reflect either Verifier error or grader error. To separate the two, we use the role-mixing pool, where every Verifier verdict is paired with the deterministic grader. The Verifier approves $49.4\%$ of submissions
that the grader rejects (Wilson $95\%$ CI $[45.9, 52.9]$, $n=1{,}083$), while the
rate of Verifier rejection on grader-passing submissions stays below $19\%$ in every cell (Figure~\ref{fig:error-analysis}-(a)). The error is therefore asymmetric, where
Verifiers more often accept failing work than reject passing work. Section~\ref{sec:errors}
analyzes these false-accept failures.

The Planner has a smaller positive effect. Adding the Planner to the No-Plan team raises mean partial score by $2.4$ points, concentrated on hard tasks where the requirements contain decision rules the Executor cannot infer. Full Team is not significantly better than Solo on average in the reference ablation. Mean team-vs-Solo
uplift is $+0.5$ points ($p=0.20$, paired bootstrap, $n=10{,}000$), and the team wins on $68$ of $155$ tasks. On \textsc{TeamBench Mini}, two stronger single-agent baselines, Solo CoT and Solo 2Pass, do not close the team gap (Appendix~\ref{app:enhanced-solo}). The average therefore hides where teams help and where they hurt.

\subsection{Where Teams Benefit}
\label{sec:capability}

\textbf{Teams help when Solo performance is low, but hurt when it is high.} We stratify the $155$-task reference ablation by per-task Solo score, with $31$
tasks per quintile. Full Team outperforms Solo by $15.7$ points in the lowest
quintile (Solo $0.00$ to $0.22$; $95\%$ CI $[5.8, 25.7]$) and by $8.8$ points
in the second quintile. In the remaining quintiles, Full Team trails Solo by
$6.8$-$10.1$ points (Figure~\ref{fig:error-analysis}-(b)).

Appendix~\ref{app:equalizer-mechanism} tests three explanations. The
pattern is not explained by time spent in Solo runs, since Solo elapsed time is
unrelated to Solo score ($r=0.00$). It is not explained by the Planner acting as a
second reasoning pass, since No-Plan and Restricted are indistinguishable on hard
tasks ($p=0.66$). It is also not explained by the amount of missing specification
alone, since the Solo--Restricted gap is negatively correlated with the Full-Team
minus Solo difference ($r=-0.45$, $p<10^{-24}$). This suggests that
team helps when Solo agent lacks enough structure to make progress. Once
Solo already makes progress, Verifier intervention can redirect or overwrite work
that would otherwise pass.

TNI classifies $15$ tasks as HIGH-TNI, $39$ as TEAM-HELPS, $62$ as NEUTRAL,
and $39$ as TEAM-HURTS. The Planner-to-Executor transfer analysis shows why
these gains are limited. Across $792$ runs over $15$ tasks, mean recall of
spec-critical tokens is $0.79$ in the Planner channel but only $0.24$ in the
Executor tool inputs and outputs. The Executor retains $0.21$ of the Planner's
spec-critical tokens on average and $0.13$ at the median. Because this measure
only counts tokens that appear in the Executor channel, it is a lower bound on how
much Planner information reaches the Executor.

\subsection{Cross-Provider Role Mixing}
\label{sec:rolemix-results}

\begin{figure}[t]
    \centering
    \includegraphics[width=\linewidth]{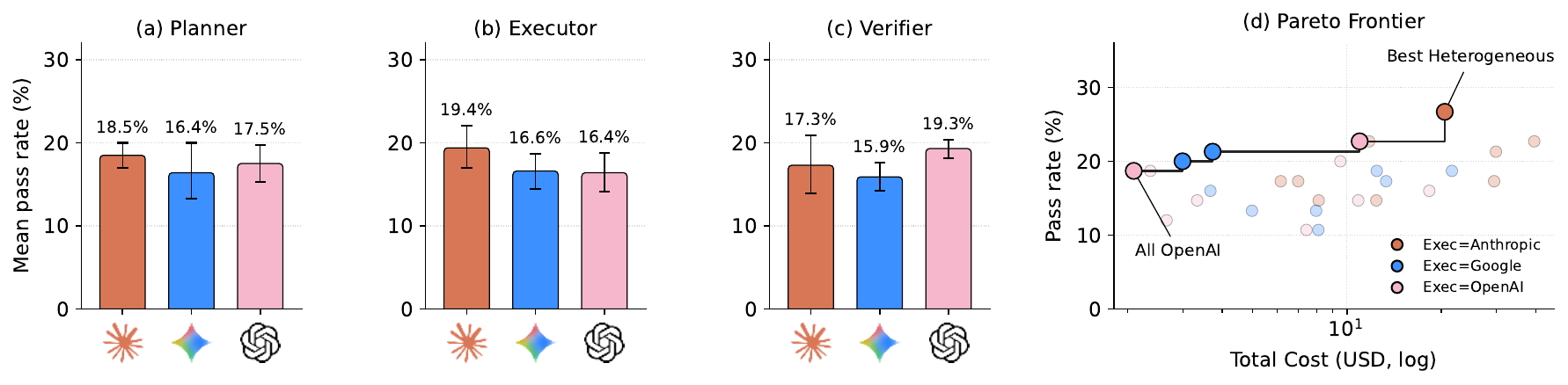}
    \caption{Cross-provider role-mixing on the $25$-task subset. \textbf{(a)-(c)} report the marginal pass rate of each provider in each role, averaged over the $9$ configurations holding that slot fixed. \textbf{(d)} plots all $27$ configurations on the cost versus pass-rate plane (log $x$). The stair-step Pareto frontier runs from POEOVO ($18.7\%$, \$2.09) to PGEAVA ($26.7\%$, \$20.52). Marker color indicates the Executor family.}
    \label{fig:rolemix}
\end{figure}

\textbf{Executor and Verifier choices matter.} 
Per-role family marginals (Figure~\ref{fig:rolemix} (a) to (c), bootstrap $95\%$ CIs in Appendix~\ref{app:rolemix-full}) show that the Anthropic Executor is about 3 points higher than the alternatives and the OpenAI Verifier has the highest Verifier marginal. The Planner confidence intervals overlap across all three families. Mixed provider teams improve the cost-performance tradeoff. Across three seeds, PGEAVA (Google Planner, Anthropic Executor, Anthropic Verifier) reaches $26.7\%$ at \$20.52, outperforming the all-Anthropic team ($22.7\%$, \$39.58) by $4$ points at roughly half the cost. POEOVA matches the all-Anthropic rate at \$10.98, and the all-OpenAI team hits $18.7\%$ at \$2.09. Fine-grained configuration ranks are unstable across seeds (Spearman $0.09$ to $0.28$, $28.9\%$ of configuration task pairs flip). We therefore focus on pooled role marginals rather than exact rank order.

\subsection{Prompt-Only vs. Enforced}
\label{sec:promptonly}

\textbf{Pass rate does not identify role behavior.} We pre-specified a 450-run study on a 25-task subset across three model families and two seeds. Prompt-only assignment shares tools and message history. Enforced assignment separates both execution environments and histories, while enforced with shared history preserves shared conversation only. After exclusions, 400 valid runs remain. We use paired McNemar tests with Holm-Bonferroni correction. No planned comparison remains significant after correction, with the strongest raw test at $p=0.038$ and $p_{\text{adj}}=0.113$. The shared-history condition is reported as inconclusive under the worst-case sensitivity analysis
(Appendix~\ref{app:enforcement-sensitivity}). Trace labels tell a different story. Enforcement reduces Verifier code-edit attempts from $256$ to $72$, while increasing executor-plans events from $261$ to $416$. These changes leave final pass
rate nearly unchanged. A pass-rate-only comparison would treat the two settings as equivalent, even though prompt-only assignment allows many more Verifier takeovers. Without enforced roles, a benchmark cannot tell whether agents coordinated or whether one model performed roles it was not assigned.

\subsection{Error Analysis}
\label{sec:errors}

\begin{figure}[t]
    \centering
    \includegraphics[width=\linewidth]{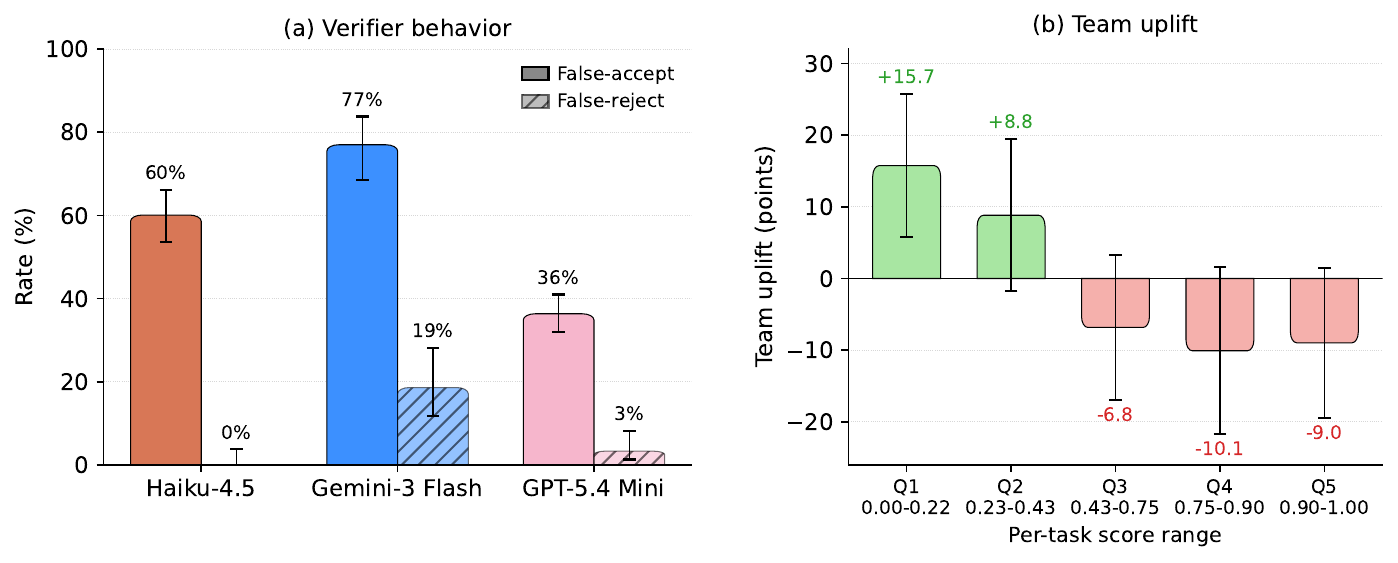}
    \caption{\textbf{Verifier errors and conditional team value.} \textbf{(a)} Verifier false-accept and false-reject rates on role-mixing runs. False-accept rates reach 60\% for Haiku-4.5, 77\% for Gemini-3 Flash, and 36\% for GPT-5.4 Mini, while false-reject rates remain below 19\%. \textbf{(b)} Mean team uplift over Solo by per-task Solo-score quintile on the 155-task reference ablation ($n=31$ per quintile). Teams help on difficult tasks but hurt on easier ones. Whiskers show 95\% confidence intervals.}
    \label{fig:error-analysis}
\end{figure}

\textbf{Verifier failure modes.} We inspect $384$ false-accept events and identify three recurring patterns (Figure~\ref{fig:error-analysis}-(a)).
Optimistic verdicts approve the submission without inspecting Executor evidence, most often for Haiku 4.5. Echo verdicts repeat the Executor's claim of completion without checking it against the requirements, most often for Gemini-3 Flash.
Verifier-rewrite failures occur when the Verifier edits workspace code-edit attempts instead of judging the Executor's output. This role takeover occurs 3.6 times more often under prompt-only assignment in the pre-specified comparison.

Counting the $942$ missing verdicts as failures lowers the effective false-accept rate from $49.4\%$ to $22.3\%$ (Table~\ref{tab:evaluator-missingness}), but leaves the conclusion unchanged. Current Verifiers are unreliable quality gates in this protocol. The audited \textsc{TeamBench-Verified} subset shows the same pattern, with a $38.7\%$ false-accept rate, suggesting that the result is not driven by one unaudited task pool. Because missing attestations are themselves verifier-side or system-side failures, the 49.4\% rate should be read as a verdict-quality estimate conditional on a verdict, while the 22.3\% sensitivity analysis is the conservative end-to-end estimate.

\textbf{Verifier errors explain why team value changes with Solo score.} Figure~\ref{fig:error-analysis}-(b) connects the Verifier false-accept result to the Solo-score stratification. In the lowest Solo-score quintile (Solo $0.00$ to $0.22$), Planner guidance offsets Verifier errors and Full Team gains $15.7$ points. In the highest quintile (Solo $\geq 0.90$), the Executor is often already close to a passing solution, and Verifier edits reduce the score by about $10$ points. This explains why the average team gain is small. Planning helps in the lowest Solo-score quintiles, while Verifier edits hurt in the highest ones.

\textbf{Small open-weight models fail mainly in tool use.} The Qwen-3 family scores at most $5.6\%$, due to invalid tool calls and context-window exhaustion, rather than task reasoning failures (Appendix~\ref{app:open-source}). At the $\leq 30$B open-weight tier,
\textsc{TeamBench} primarily measures tool-call reliability rather than coordination.

\subsection{Human Study}
\label{sec:human-study}

Appendix~\ref{app:human-platform} reports a human study run under settings that mirror the agent experiments. \textbf{Solo} is a single human with full access. \textbf{Hybrid} is one human paired with two agents. \textbf{Team} is three humans under the same role separation as the agent harness. The study covers $40$ completed survey-confirmed sessions across $21$ tasks from $18$ distinct humans (Solo $n{=}12$, Hybrid $n{=}17$, Team $n{=}11$). $13$ of the $20$ stratified target tasks have at least one session. The pilot is covered by an MIT COUHES exempt determination under 45 CFR 46.104(d)(2), Exempt Category 2.

\textbf{The value of the verifier.} Across $32$ role-level surveys from $11$ Team sessions, the three CATME \citep{o2024objective} ratings cluster tightly. Verifier value is $3.75$, Executor efficiency is $3.72$, and early planning is $3.69$ on a $1$ to $5$ scale. In the Solo counterfactual survey ($n=12$), participants rate per-role teammates similarly ($3.00$ to $3.17$) and rate the time-only counterfactual lowest at $2.17$, indicating that adding any teammate is perceived as more helpful than additional alone-time. The Verifier role is therefore not perceived as low-value in either instrument, in contrast to the role where LLMs fail most often in our agent experiments (\S\ref{sec:errors}). The contrast suggests that the Verifier slot is not inherently low value, but that current Verifiers struggle to execute it reliably in this protocol.

\textbf{Time pressure and missing information lead the failure factors.} Figure~\ref{fig:human-pilot}-(c) plots the primary failure factors from the $32$ Team surveys. Time pressure is the top endorsement ($17$ of $32$), followed by information missing across roles ($14$). The next four are weak or late planning ($7$), unclear communication ($7$), implementation difficulty ($6$), and other ($4$), with missed verification tied at $4$ ($12.5\%$). Two observations follow. First, humans running the same single-pass file-based protocol cite the same architectural property the benchmark imposes on agents, namely information structurally split between the specification and the workspace, as their dominant non-time failure cause. Second, missed verification is rarely selected by humans, which contrasts with the $49\%$ Verifier false-accept rate in the agent experiments (\S\ref{sec:errors}). In this study, human participants rarely selected missed verification as the primary failure factor, whereas Verifiers often approved grader-failing work.

\textbf{Three modes show different collaboration patterns.} \emph{Solo} sessions are exploratory, with median $11$ minutes wall-clock and median $6$ explicit actions per session. \emph{Hybrid} sessions collapse to a near-instant approve-and-submit pattern, with median $3$ minutes and only $2$ explicit actions per session, while $16$ of $17$ Hybrid Verifiers self-attested pass. \emph{Team} sessions instead spread effort across roles, with median $26$ minutes and median $39$ explicit actions per session. Per-role chat turns (Figure~\ref{fig:human-pilot}-(b)) place the Executor highest (median $10$), the Planner middle (median $8$), and the Verifier lowest (median $5$), even though the Verifier role receives the highest perceived value.

%% ============================================================
%% 5. DISCUSSION
%% ============================================================
\section{Discussion}
\label{sec:discussion}

\textbf{Pass rate alone is not enough for agent evaluation.} The $40.5\%$ Full Team pass rate in the prompt-only versus enforced study
(Table~\ref{tab:promptonly}) hides two effects that matter for coordination. Enforcement reduces Verifier code-edit attempts by 3.6 times, and the Full-Team minus Solo gap changes by $24.7$ points between the lowest and highest Solo-score quintiles. Reporting only final pass rate would miss both. We suggest two additions to multi-agent benchmark reports. First, reports should include role-violation rates from per-turn traces, so that teams with the same pass rate but different role behavior are not treated as equivalent. Second, reports should stratify team value by Solo score, so that where teams help and where they hurt is visible rather than averaged into a $+0.5$ point
mean. \textsc{TeamBench} reports both quantities and the per-turn rubric needed to compute role violations.

\textbf{LLM judges may inherit the same false-accept problem.} In the role-mixing runs, Verifiers accept grader-failing submissions at rates from
$36\%$ for GPT-5.4 Mini to $77\%$ for Gemini-3 Flash, with a pooled rate of $49.4\%$ (Figure~\ref{fig:error-analysis}-(a)). This matters for multi-agent benchmarks that use an LLM judge to mark milestones or task completion, including
the MultiAgentBench Coordination Score~\citep{zhu2025multiagentbench}. Such
metrics can overcredit outputs that satisfy the judge while failing deterministic checks. A simple safeguard is to anchor scores to deterministic graders or human-audited outcomes, and to report judge-grader disagreement. The $49\%$
rate is a property of our sampled role-mixing distribution, not of any single model. Future systems may reduce it with tool-assisted checking or attestations that require evidence.

\textbf{Matched human runs make the process signal visible.} The human study under the same role boundaries gives three signals that agent
traces alone cannot provide (Appendix~\ref{app:human-platform}).
First, humans report missing information across roles as their main non-time barrier, matching the split between specification access and workspace access imposed by the benchmark. Second, Hybrid sessions often reduce to quick approval, with
median $3$ minutes per session and $94.1\%$ self-attestation against $79\%$ structural-grader partial mean. This mirrors the over-acceptance seen in Verifiers and suggests that human-in-the-loop benchmarks should log override behavior.
Third, the Verifier receives the highest stated value but the lowest activity count, even though Verifiers fail most often in the agent experiments. Together, these observations show that pass rate alone is not a sufficient signal for
multi-agent evaluation, whether the final decision is made by a model or by a human in the loop. We release the benchmark and human-baseline platform together so that future systems can be evaluated under the same single-pass file-based protocol.

\textbf{Cross-provider mixing changes the cost-performance frontier.} On the $25$-task grid, the Anthropic-only team reaches $22.7\%$ at \$39.58.
PGEAVA reaches $26.7\%$ at \$20.52, POEOVA matches the Anthropic-only rate at \$10.98, and the all-OpenAI team reaches $18.7\%$ at \$2.09 (Appendix~\ref{app:rolemix-full}). No single provider dominates the Pareto frontier across the $27$ configurations. The per-role
marginals suggest that Executor and Verifier choice matters more than Planner choice. This means that single-provider leaderboards may miss cheaper mixed-provider configurations with similar or better pass rates. We release the full $27$-cell grid
and cost ledger to support cost-aware multi-agent comparisons.

\textbf{Relation to prior work and remaining scope.} The Solo-score pattern in Section~\ref{sec:capability} matches Steiner's
prediction~\citep{steiner1972group} that disjunctive tasks yield little team gain
over a capable individual. The Executor retains only $0.21$ of the Planner's spec-critical tokens on average, consistent with transactive memory findings on newly formed teams~\citep{wegner1987transactive,lewis2003measuring}. The human study points in the same direction: participants rate the Verifier as useful, while
LLM Verifiers fail frequently in the agent experiments. Compared with concurrent
diagnostics that taxonomize failures~\citep{cemri2025whyfail}, probe collective
reasoning~\citep{li2025hiddenbench}, or formalize benchmark validity~\citep{zhu2025abc},
\textsc{TeamBench} adds a controlled comparison of prompt-assigned and enforced roles. The current harness uses capped turns and a file-based workflow. It does not test multi-round dialogue, dynamic role assignment, or within-provider model-size scaling. Verifier false acceptance should be a priority for future agent systems. Enforced role separation makes that target measurable. A stronger Verifier should reject failing outputs without taking over the Executor role.

%% ============================================================
%% 5. CONCLUSION
%% ============================================================
\section{Conclusion}
\label{sec:conclusion}

\textsc{TeamBench} enforces role separation with sandbox permissions and evaluates the resulting teams across 13 models. We find that teams rarely outperform single agents on average, Verifiers falsely accept many failing submissions, team gains depend strongly on Solo capability, and a mixed provider team can match the strongest single provider team at roughly half the cost. Prompt-only and enforced assignments reach statistically indistinguishable pass rates, but prompt only runs produce 3.6 times more Verifier code-edit attempts. We release the benchmark, harness, and human study platform so that agent systems can be evaluated not only by final pass rate, but by whether each added agent performs distinct work.

%% ============================================================
%% REFERENCES
%% ============================================================
\bibliographystyle{unsrtnat}
\bibliography{references}

%% ============================================================
%% APPENDIX
%% ============================================================
\newpage

\appendix

\section{Benchmark Construction}
\label{app:construction}

\subsection{Role permissions}
\label{app:role-permissions}

\begin{table}[h]
\centering
\caption{Role permissions enforced by Docker bind mounts. No single role has simultaneous access to the full specification, workspace write access, and attestation write access.}
\label{tab:roles}
\small
\begin{tabular}{lccc}
\toprule
 & \textbf{Planner} & \textbf{Executor} & \textbf{Verifier} \\
\midrule
Read \texttt{spec.md}                      & \checkmark & --        & \checkmark            \\
Read \texttt{brief.md}                     & \checkmark & \checkmark & --                    \\
Read \texttt{workspace/}                   & --        & \checkmark & \checkmark (read-only) \\
Write \texttt{workspace/}                  & --        & \checkmark & --                    \\
Read \texttt{messages/}                    & \checkmark & \checkmark & \checkmark            \\
Write \texttt{messages/}                   & \checkmark & \checkmark & \checkmark                    \\
Read \texttt{reports/} (Executor logs)     & --        & \checkmark & \checkmark (read-only) \\
Write \texttt{reports/}                    & --        & \checkmark & --                    \\
Write \texttt{attestation.json}            & --        & --        & \checkmark            \\
Execute commands                           & --        & \checkmark & --                    \\
\bottomrule
\end{tabular}
\end{table}

\subsection{Task construction}
\label{app:task-construction}

\paragraph{Originally-authored templates.}
Information asymmetry must be introduced deliberately. In benchmarks derived from real GitHub issues, the issue description typically contains enough context for any single agent to proceed, rendering the Planner redundant by construction. Our 161 originally-authored templates enforce asymmetry at the authoring stage by placing critical constraints exclusively in \texttt{spec.md}, absent from the brief and workspace.

\paragraph{Real GitHub bug reports.}
The benchmark includes 11 hand-curated GitHub bug tasks (GH1 through GH11) drawn from Flask, Click, httpx, Requests, Pydantic, Django, pytest, FastAPI, SQLAlchemy, Celery, and Werkzeug, plus 639 additional templates from a broader scrape across active open-source repositories such as NumPy, SciPy, Keras, and spaCy. The hand-curated set includes the full issue discussion in the specification and a user-facing symptom in the brief.

\paragraph{Task file layout.}
Each task includes a specification (\texttt{spec.md}) with all requirements and edge cases, a brief (\texttt{brief.md}) for the Executor, a workspace with initial code, a setup script, and a deterministic grader (\texttt{grade.sh}) that produces a partial score in $[0, 1]$.

\paragraph{Evaluation subsets.}
The $28$-task \textsc{TeamBench-Mini} subset is retained for backward-compatible quick validation. The new stratified \textsc{TeamBench} leaderboard (Appendix~\ref{app:lb90-construction}) supersedes Mini as the primary cross-model evaluation set.

\subsection{Related work}
\label{app:related-extended}

\paragraph{Single-agent and agent-scaffolding benchmarks.}
SWE-Bench~\citep{jimenez2024swebench}, Terminal-Bench~\citep{terminalbench2026}, MLE-Bench~\citep{chan2024mlebench}, DevBench~\citep{li2024devbench}, GAIA~\citep{mialon2023gaia}, BrowseComp~\citep{browsecomp2025}, and AgentBench~\citep{liu2024agentbench} drive progress on single-agent capability across software engineering, terminal use, ML engineering, multi-step reasoning, and open-web browsing. SWE-agent~\citep{yang2024sweagent} and OpenHands~\citep{wang2024opendevin} package the scaffolding side. None of these benchmarks include ablation conditions that measure multi-agent coordination directly. LiveCodeBench~\citep{jain2024livecodebench} uses temporal holdouts for contamination resistance. Our parameterized generators produce arbitrary unseen seeds at evaluation time, decoupling contamination resistance from the publication date.

\paragraph{Multi-agent frameworks and benchmarks.}
MetaGPT~\citep{hong2024metagpt}, AutoGen~\citep{wu2023autogen}, ChatDev~\citep{qian2024chatdev}, CAMEL~\citep{li2023camel}, multi-agent debate~\citep{du2024debate}, and GPTSwarm~\citep{zhuge2024gptswarm} assign roles through prompt instructions, which permits a single dominant agent to absorb every role and conflates coordination with prompt compliance. Our container-based harness enforces role separation directly rather than relying on prompt instructions. Table~\ref{tab:benchmark-comparison} compares \textsc{TeamBench} against this group across eight design axes. MultiAgentBench~\citep{zhu2025multiagentbench} is the closest concurrent benchmark in spirit, evaluating six interactive scenarios across four communication topologies (star, chain, tree, graph) with an LLM-based milestone detector that yields Task and Coordination Scores. Two design choices distinguish our work. The first is to fix a single Planner-Executor-Verifier topology and vary enforcement rather than vary topology under prompt-assigned roles, which isolates whether coordination gains come from the team structure or from prompt compliance, an axis MultiAgentBench does not isolate. The second is to evaluate coordination against a deterministic grader rather than an LLM judge on milestone completion. Our Verifier-false-acceptance measurement ($49\%$ on grader-failing runs) raises the possibility that LLM-judge coordination scores may over-credit teams unless calibrated against deterministic or human-audited outcomes, which we view this as a calibration check rather than a competing benchmark.

\paragraph{Concurrent multi-agent diagnostics.}
MAST~\citep{cemri2025whyfail} extracts a 14-mode failure taxonomy from 1{,}600 traces across seven prompt-orchestrated frameworks and concludes that prompted teams underperform single agents on a variety of tasks. The controlled experimental setting we report generates the kind of traces such a taxonomy needs, and isolates how much failure attribution depends on prompt-only role assignment. HiddenBench~\citep{li2025hiddenbench} ports the social-psychology hidden-profile paradigm~\citep{stasser1985pooling} to multi-agent LLMs on 65 abstract-decision tasks. We measure coordination on production-grade software and data-analysis tasks instead. The Agentic Benchmark Checklist~\citep{zhu2025abc} documents validity failures in agent benchmarks and proposes task-validity, outcome-validity, and reporting standards that we adopt by separating the deterministic grader from the Verifier attestation, by reporting the Verifier false-acceptance rate against the grader, and by reporting per-seed flip rates rather than headline ranks alone. Anthropic's multi-agent retrospective~\citep{anthropic2025multiagent} reports that subagents in coding workloads spend more tokens on coordination than on actual work, matching our finding that the average team gain is small and that the Verifier is the binding constraint.

\paragraph{Connections to human teamwork research.}
The capability-floor pattern in Section~\ref{sec:capability} aligns with Steiner's disjunctive-task prediction~\citep{steiner1972group}. The Planner-to-Executor relay fidelity of $0.21$ aligns with the transactive-memory line~\citep{wegner1987transactive, lewis2003measuring} on fresh-strangers teams. Hackman~\citep{hackman1987design} treats process measurement and communication quality as prerequisites for explaining team performance differences, and the released traces give downstream work a reusable process-measurement substrate. The Discussion expands on the implications.

\section{Human Study}
\label{app:human-platform}

\begin{figure}[ht!]
    \centering
    \includegraphics[width=\linewidth]{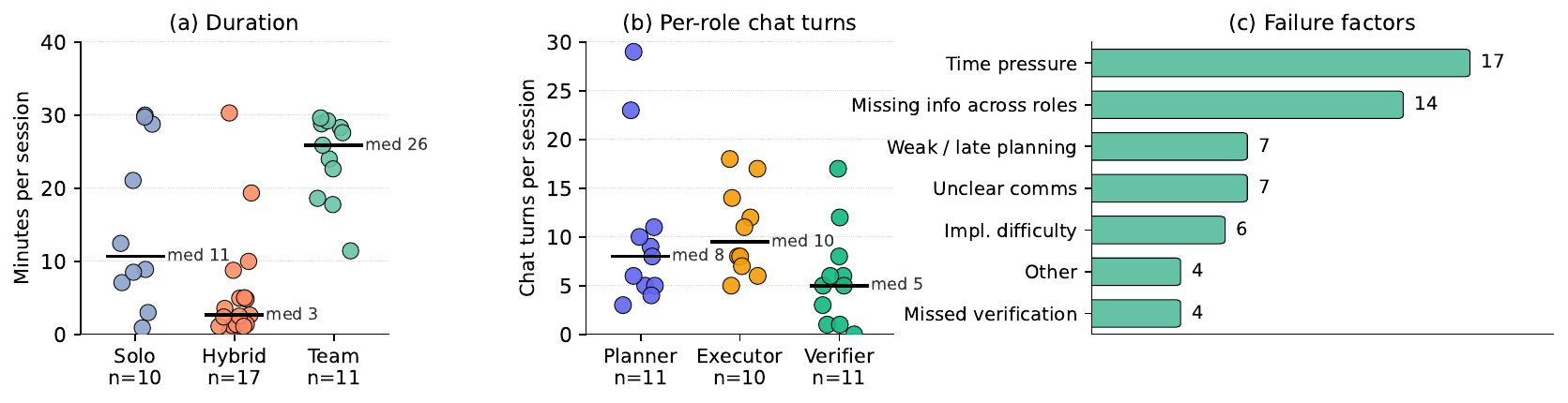}
    \caption{\textbf{Human study results.} Each colored dot is one session, and the short black horizontal bar marks the per-mode median. \textbf{(a)} Duration per session by mode, in minutes on a linear scale, capped at $40$ minutes ($n=10$ Solo, $n=17$ Hybrid, $n=11$ Team within the cap; two Solo sessions were excluded since they exceeded the cap). \textbf{(b)} Per-role chat turns in Team mode, counting only messages a participant sent in the chat panel. The per-session total is reported in the body text. \textbf{(c)} Failure factors from Team-mode participants on the post-task survey, multi-select per survey across $32$ role-level responses from $11$ Team sessions.}
    \label{fig:human-pilot}
\end{figure}

\begin{figure}[h]
    \centering
    \includegraphics[width=0.8\linewidth]{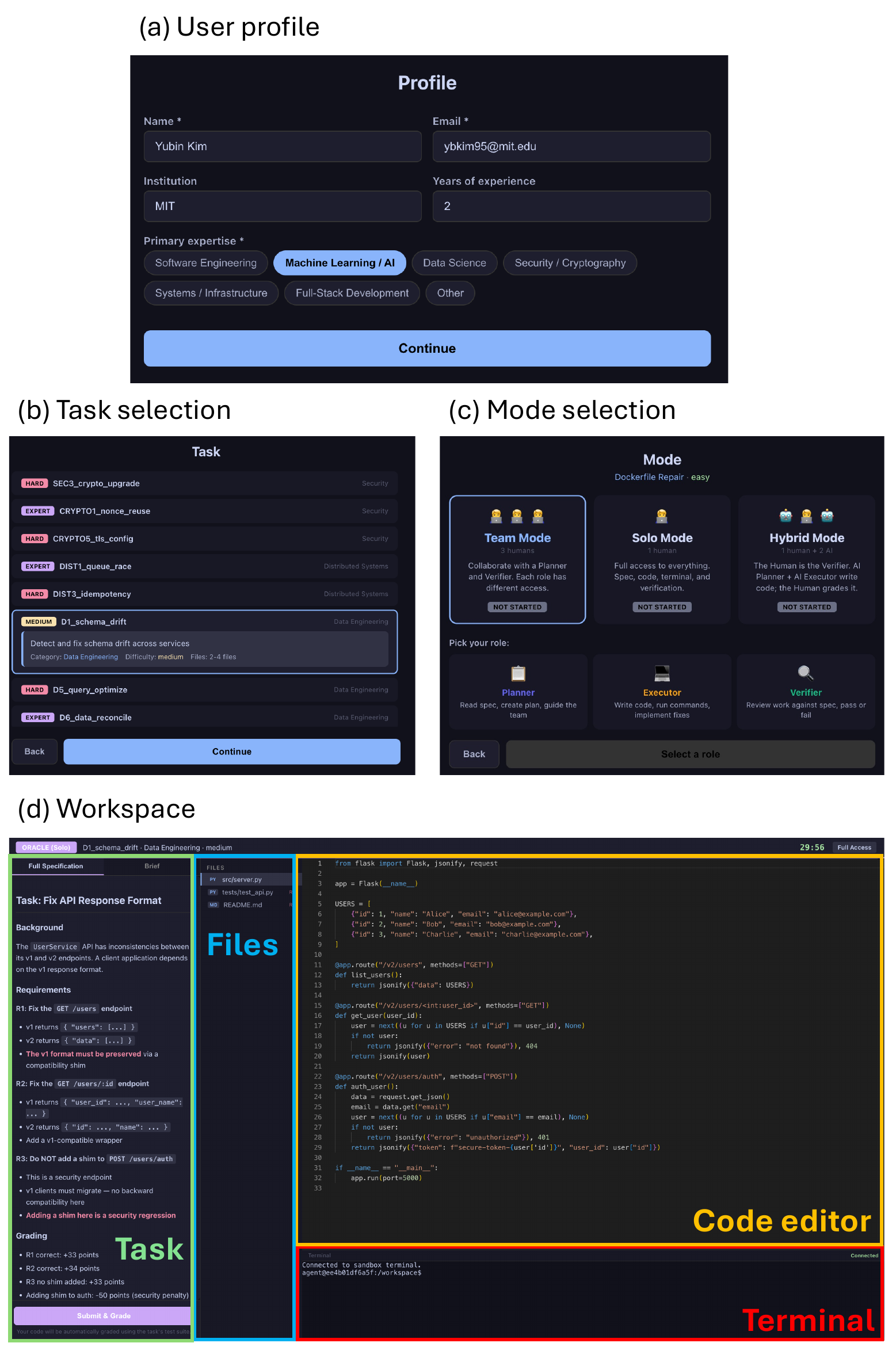}
    \caption{\textbf{Human study platform} (\url{https://teambench.github.io/human-eval/}). Participants enter a profile (a), pick a task from the stratified human-eval subset (b), select Team or Solo Mode with their role (c), and work in a sandboxed workspace with the same graders and role-based file access that agents use (d). The platform issues the same attestation file that closes the agent run.}
    \label{fig:human-platform}
\end{figure}

We deploy a web-based collection platform that runs \textsc{TeamBench} tasks for human participants under the same role separation. Figure~\ref{fig:human-platform} shows the four screens of the flow. Participants sign in with their institution and primary expertise (a), pick a task from the stratified human-eval subset (b), choose Team Mode or Solo Mode and their role (c), and then work in a workspace view that mirrors the agent harness with identical file scoping, grading, and attestation flow (d). The same Docker grading script scores human and agent submissions, which allows direct human-versus-agent comparison on matched tasks. The current submission reports a pilot and releases the platform for larger follow-up studies. The platform logs participant behavior by task and role, and can be reused by other groups for matched human comparisons.

\subsection{Pilot Coverage}
\label{app:human-coverage}

Coverage as of submission, derived from the released collection backend. A row is counted only if the session reached the \texttt{completed} phase, the participant submitted a survey for the participant's role, and the identity passes a dev-pollution filter (real-shaped email, name length $\geq 2$, no test/admin/probe pattern). 

% Two notes apply to the identity counts: a participant whose institutional email is placeholder-shaped is genuine and not filtered, and one participant used two distinct institutional emails, so the $19$ unique emails below correspond to $18$ distinct humans.

\begin{table}[h]
\centering
\small
\caption{Human baseline pilot coverage as of submission, restricted to sessions that reached the completed phase with a submitted survey.}
\label{tab:human-coverage}
\begin{tabular}{lrrrr}
\toprule
mode & unique humans & unique emails & sessions & distinct task ids \\
\midrule
Solo   & 6$^{\dagger}$ &  7 & 12 & 10 \\
Hybrid & 10            & 10 & 17 & 11 \\
Team   & 9$^{\dagger}$ & 10 & 11 &  8 \\
\midrule
total  & 18$^{\dagger}$ & 19 & 40 & 21$^{\ddagger}$ \\
\bottomrule
\end{tabular}
\\[2pt]
\footnotesize
$^{\dagger}$ Email-uniqueness lists Solo as $7$ and Team as $10$ because one participant used two emails in each mode. De-duplicated by person, Solo is $6$ humans, Team is $9$, and the total is $18$. $^{\ddagger}$ $21$ distinct task ids across all modes, of which $13$ overlap with the $20$-task stratified target subset.
\end{table}

Of the $20$ stratified target tasks, Solo has any data on $5$ (\texttt{API1\_version\_compat}, \texttt{CR4\_api\_review}, \texttt{D6\_data\_reconcile}, \texttt{TEST3\_integration}, \texttt{TRAP1\_spec\_conflict}), Hybrid on $5$ (\texttt{DIST1\_queue\_race}, \texttt{GH1002\_scipy\_24753}, \texttt{O6\_perf\_tuning}, \texttt{RDS10\_survey\_analysis}, \texttt{RDS13\_smote\_leakage}), and Team on $5$ (\texttt{CR4\_api\_review}, \texttt{IR2\_misinformation\_trap}, \texttt{LH2\_budgeted\_workflow}, \texttt{PIPE2\_data\_pipeline}, \texttt{TEST3\_integration}).

\subsection{Findings}
\label{app:human-measurement}

\paragraph{Hybrid wall-clock distribution.} Of the $17$ Hybrid sessions with a recorded duration, $1$ finished under $60$\,s, $8$ between $1$ and $3$ minutes, $5$ between $3$ and $10$ minutes, $2$ between $10$ and $30$ minutes, and $1$ between $30$ and $60$ minutes. The $12$ Solo sessions distribute much more evenly across the same buckets: $1, 1, 3, 4, 2$, with one additional Solo sessions above $60$ minutes excluded from the Figure~\ref{fig:human-pilot}(a) duration panel as an inactive browser session. The collapse of the Hybrid bottom buckets ($9$ of $17$ under $3$ minutes), combined with the $16 / 17$ self-attestation rate against the deterministic-grader partial mean of $0.79$ (\S\ref{sec:human-study}), is the basis for the grader-override measurement finding.

\paragraph{Survey instruments.} Solo participants complete the \textsc{TeamBench-Solo-Reflection} instrument: an attention check, a counterfactual team-value Likert (per-role and time-only items), and an open-ended collaboration-challenge prompt. Hybrid participants complete a \textsc{Hybrid-AI-Teammate} instrument that asks how useful and trustworthy the AI Planner and Executor were, how often the participant overrode the grader (\texttt{overrode\_grader}, $1$=never, $5$=always), and the perceived value of the Verifier role. Team participants complete a \textsc{CATME + TeamBench Coordination} instrument with role-separation, verifier-value, communication-overhead, early-plan, executor-efficiency, and information-held-by-other items, plus the same attention check.

\paragraph{Self-attestation systematically overstates completion.} The methodology, exclusions, and per-session results are in Appendix~\ref{app:human-regrade}. We report the structural-checks partial score as the primary metric since the binary all-checks-pass column is sensitive to a single \texttt{pytest} invocation that depends on container-level environment parity which our offline pipeline does not fully reproduce. Hybrid self-attested $16 / 17$ pass and the grader confirms a structural mean partial of $0.79$ (median $0.86$). Team self-attested $11 / 11$ pass and the grader confirms a structural mean partial of $0.60$ (median $0.83$, with $4$ of $11$ sessions reaching $\geq 0.86$ and the strongest at $0.94$ on \texttt{CR4\_api\_review}). Solo verdicts are not written by the current platform, but the $12$ Solo sessions that re-graded yield a structural mean partial of $0.80$ (median $0.90$). Team-mode submissions are therefore produced at near-completion quality, while the human Verifier role accepts them as fully complete every time. The Hybrid post-task survey corroborates the override pattern. The mean self-reported \texttt{overrode\_grader} score is $2.41$ on a $1$ to $5$ scale, and $9$ of $17$ sessions report $\geq 3$. We release the platform, survey instruments, and pre-specified analysis plan, so other groups can run matched human comparisons at scale.

\paragraph{Pilot survey aggregates (descriptive only, $N$ small).}

\begin{table}[h]
\centering
\small
\caption{Pilot survey aggregates by mode. Likert items are $1$ to $5$, mean reported with sample size in parentheses. For Solo and Hybrid, $n$ counts sessions ($n=12$ Solo, $n=17$ Hybrid). For Team, $n=32$ counts \emph{role-level survey responses} pooled across team members from $11$ Team sessions. Means within $\sim 0.1$ of each other should be read as clustered, not ranked.}
\label{tab:human-pilot-surveys}
\begin{tabular}{lll}
\toprule
mode & item & mean ($n$) \\
\midrule
Solo (counterfactual)
& would a Planner teammate help (\texttt{cf\_planner})    & $3.17$ ($12$) \\
& would a Verifier teammate help (\texttt{cf\_verifier})  & $3.08$ ($12$) \\
& would an Executor teammate help (\texttt{cf\_executor}) & $3.00$ ($12$) \\
& would domain expertise help (\texttt{cf\_domain})       & $3.00$ ($12$) \\
& would more time alone help (\texttt{cf\_time\_only})    & $2.17$ ($12$) \\
\midrule
Hybrid (AI teammate)
& AI Planner useful (\texttt{ai\_planner\_useful})        & $3.76$ ($17$) \\
& AI Planner trust (\texttt{ai\_planner\_trust})          & $3.71$ ($17$) \\
& AI Executor quality (\texttt{ai\_executor\_quality})    & $3.59$ ($17$) \\
& AI Executor trust (\texttt{ai\_executor\_trust})        & $3.71$ ($17$) \\
& Verifier role value (\texttt{verifier\_role\_value})    & $3.29$ ($17$) \\
& self-reported grader-override (\texttt{overrode\_grader}) & $2.41$ ($17$) \\
\midrule
Team (coordination)
& verifier value                                          & $3.75$ ($32$) \\
& executor efficiency                                     & $3.72$ ($32$) \\
& early plan                                              & $3.69$ ($32$) \\
& role separation helped                                  & $3.44$ ($32$) \\
& information held by other                               & $3.38$ ($32$) \\
& communication overhead                                  & $3.25$ ($32$) \\
\midrule
Team (counterfactual ``stronger X'' items)
& stronger Executor would change outcome                  & $3.72$ ($32$) \\
& stronger Verifier would change outcome                  & $3.53$ ($32$) \\
& stronger Planner would change outcome                   & $3.44$ ($32$) \\
\bottomrule
\end{tabular}
\end{table}

\paragraph{Team-mode primary failure factors.} The Team coordination instrument includes a structured multi-select item (max $3$ selections) asking which factor most affected the outcome on this task. Across the $32$ team-role surveys, endorsement counts are: \texttt{time\_pressure} $17$, \texttt{missing\_info\_across\_roles} $14$, \texttt{weak\_or\_late\_planning} $7$, \texttt{unclear\_communication} $7$, \texttt{implementation\_difficulty} $6$, \texttt{other} $4$, \texttt{missed\_verification} $4$.

\subsection{Re-graded human outcomes}
\label{app:human-regrade}

The verdict field stored against each session is the participant's self-attestation written from the verdict UI, not a deterministic-grader run. We re-grade each session offline by replaying the participant's final workspace snapshot through the same shell-script grader used by the agent harness. The pipeline is deterministic and reproducible from the released collection backend export.

For each session that reached the completed phase with a submitted survey, we replay the final workspace snapshot under the same deterministic shell-script grader used by the agent harness, materializing each file into a fresh temporary workspace and using the generator's seed-0 expected outputs. Support files outside the canonical workspace are written best-effort.

\begin{table}[h]                             
\centering                                   
\small                                       
\caption{Re-graded human outcomes versus self-attestation. Self-attest counts the participant's pass verdict written to the attestation file. Grader counts a pass under the deterministic shell-script grader after replaying the final workspace snapshot. All 40 completed sessions regrade cleanly under the offline pipeline.}                         
  \label{tab:human-regrade}               
  \begin{tabular}{lrrrrrr}                    \toprule                                    mode & $n$ & self-attest pass & binary grader pass$^{\dagger}$ & mean partial & median partial & top partial \\   
  \midrule                                    Solo   & $12$ & not written      & $1$ ($8.3\%$)  & $0.80$ & $0.90$ & $1.00$ \\    Hybrid & $17$ & $16$ ($94.1\%$)  & $4$ ($23.5\%$) & $0.79$ & $0.86$ & $1.00$ \\    Team   & $11$ & $11$ ($100\%$)   & $0$ ($0\%$)$^{\ddagger}$ & $0.60$ & $0.83$ & $0.94$ \\                                   \bottomrule                                \end{tabular}
\end{table}  
 
\noindent$^{\dagger}$Partial score is the fraction of grader checks satisfied and is robust to the offline-pipeline limitation flagged below. The binary-pass column requires every check including a containerized \texttt{pytest} invocation that our offline pipeline cannot fully reproduce. $^{\ddagger}$Team binary $0/10$ is therefore a lower bound. Several Team sessions reach $0.86$ to $0.88$ partial (one failing test away from binary pass).

The interpretation is the self-attestation gap. Team mode produces structurally near-complete artifacts (median partial $0.83$, top $0.94$) and the human Verifier role accepts every one of those grader-failing submissions ($10 / 10$). On the role-mixing pool, the Verifier false-accept rate is $49\%$ on grader-failing runs (\S\ref{sec:errors}). The pilot is small and the Team-mode tasks are hard-skewed ($5$ of $7$ unique tasks are hard or expert), so we report this as a directional confirmation rather than a precise human-vs-LLM ranking. The eligibility filter and minimum-cell rules used for the reported aggregates are stated in \S\ref{app:human-prereg}.

\begin{table}[t]
\centering
\caption{pre-specified prompt-only versus structural-enforcement comparison. The primary prompt-only vs enforced contrast has $148$ paired observations. The shared-history condition has lower coverage due to infrastructure exclusions and is treated as inconclusive. Per-run violation rate is the per-run mean fraction of turns flagged by a deterministic role-compliance rubric. McNemar tests are paired at the (model, task, seed) level with Holm-Bonferroni correction.}
\label{tab:promptonly}
\small
\setlength{\tabcolsep}{6pt}
\begin{tabular}{lrrr}
\toprule
Condition & Pass rate (\%) [95\% CI] & Per-run violation rate (\%) [95\% CI] & n \\
\midrule
\texttt{prompt\_only}              & 42.7 [34.7, 50.0] & 6.4 [5.3, 7.6]  & 150 \\
\texttt{enforced}                  & 40.5 [32.4, 48.6] & 6.2 [5.3, 7.3]  & 148 \\
\texttt{enforced\_shared\_history} & 48.0 [38.2, 57.8] & 8.9 [7.6, 10.3] & 102 \\
\midrule
\multicolumn{4}{l}{\textit{pre-specified McNemar tests (Holm-Bonferroni adjusted)}} \\
\multicolumn{2}{l}{T1: compliance, prompt-only vs enforced} & $p_{\text{adj}}=0.113$ & 148 pairs \\
\multicolumn{2}{l}{T2: outcome, prompt-only vs enforced} & $p_{\text{adj}}=0.907$ & 148 pairs \\
\multicolumn{2}{l}{T3: outcome, shared-history vs enforced} & $p_{\text{adj}}=0.907$ & 100 pairs \\
\bottomrule
\end{tabular}
\end{table}

\subsection{Analysis Protocol}
\label{app:human-prereg}

The analysis pipeline applies the following decisions, fixed before the human dataset was re-opened for analysis.

\begin{enumerate}[leftmargin=*]
\item \textbf{Eligibility filter.} A Hybrid session is analyzable iff \texttt{overrode\_grader} $\leq 2$ on the 1 to 5 scale and wall-clock $\geq 5$ minutes. Sessions failing either criterion are reported but excluded from outcome comparisons. Solo and Team sessions have no analogous eligibility filter beyond the existing attention check.
\item \textbf{Minimum cell.} A per-task pass rate is reported only at $\geq 10$ participant-sessions in the relevant cell, and an aggregate is reported only if $\geq 8$ tasks satisfy the per-task minimum-cell rule.
\item \textbf{Pairing.} Solo-vs-LLM-Solo and Team-vs-LLM-Team comparisons use McNemar's exact test on (task id, seed) pairs, with Holm-Bonferroni adjustment across the three pairings (Solo, Hybrid, Team).
\item \textbf{Effect-size focus.} Headline numbers are paired differences in pass rate with Wilson 95\% confidence intervals. Significance tests are secondary. We do not chase $p < .05$ thresholds.
\item \textbf{Open-ended responses} are summarized by two annotators using the same role-collapse rubric as Section~\ref{sec:ref-results}, and inter-annotator agreement is reported as Cohen's $\kappa$.
\item \textbf{Demographics, recruitment, and IRB.} As specified in the public protocol document accompanying the platform release.
\end{enumerate}

Stop condition for collection. The target is $\geq 10$ sessions per mode on each of the 20 stratified target tasks, with the eligibility filter applied.

\section{Leaderboard}
\label{app:leaderboard-eval}

\subsection{Studies and counts}
\label{app:study-counts}

\begin{table}[h]
\centering
\caption{The source of truth for counts and studies in the paper.}
\label{tab:studies}
\small
\setlength{\tabcolsep}{5pt}
\begin{tabular}{lrl}
\toprule
Quantity & Count & Used in \\
\midrule
Templates                                          & $851$ & full pool \\
Seeded instances                                   & $931$ & full pool \\
Base / refined categories                          & $19$ / $21$ & taxonomy \\
Leaderboard subset                                 & $90$  & leaderboard ($5$ conditions, $13$ models) \\
\textsc{TeamBench-Verified} subset                 & $57$  & audited subset for grader-sensitive analyses \\
Reference ablation pool                            & $155$ & $5$-condition runs on gemini-3-flash \\
Cross-provider grid                                & $25$  & $27$ configs $\times$ $3$ seeds, $3$ families \\
pre-specified enforcement                         & $25$  & $3$ conditions $\times$ $2$ seeds, $3$ families \\
Models evaluated in the leaderboard     & $13$ & panel includes models with valid runs \\
\bottomrule
\end{tabular}
\end{table}

\subsection{Leaderboard construction}
\label{app:lb90-construction}

The \textsc{TeamBench} leaderboard subset was selected by stratified sampling within each of the $21$ refined categories. Quotas were set proportional to category size in the full pool, rounded to whole tasks, with a floor of one task per category. Within each category, tasks were sampled with priority on (a) presence of complete reference ablation data, then (b) difficulty-mix balance, then (c) authoring date as tie-breaker. The selection script and the selection JSON ship in the public release. Table~\ref{tab:lb90-detail} gives the per-category quota and difficulty mix. The ``difficulty mix'' columns in Tables~\ref{tab:lb90}~and~\ref{tab:lb90-detail} use the per-template author label; the objective grader-check rubric in Figure~\ref{fig:full-pool-pie}(c) covers the full 851-template pool.

\begin{table}[t]
\centering
\caption{Task distribution. The top block lists the difficulty breakdown for the 153 originally-authored templates with complete 5-condition ablation metadata. Two extension tasks with later-added ablation data join this core to form the 155-task reference ablation pool used in \S\ref{sec:capability}. The lower block lists the remaining 698 templates by origin (650 GitHub bug reports, 30 real data-science, 10 real incident response, 8 originally-authored without ablation yet) for a total of 851 unique templates and 931 seeded evaluation instances. Difficulty in this table uses the per-template author label; the objective grader-check rubric in Figure~\ref{fig:full-pool-pie}(c) covers the full 851-template pool.}
\label{tab:distribution}
\small
\begin{tabular}{lccccr}
\toprule
Category (Original) & Easy & Medium & Hard & Expert & Total \\
\midrule
Security (SEC/CRYPTO)            & --  & 1   & 10  & 4   & 15 \\
Data Engineering (D/SQL)         & 1   & 4   & 5   & 1   & 11 \\
Incident Response (INC)          & --  & 1   & 9   & 1   & 11 \\
Software Eng. (SWE/GO/JS)        & 1   & 1   & 8   & 1   & 11 \\
Testing (TEST)                   & 1   & 2   & 6   & --  & 9 \\
Multi-language (MULTI)           & --  & --  & 9   & --  & 9 \\
Operations (OPS)                 & 1   & 2   & 5   & 1   & 9 \\
Pipeline / Integration (PIPE/INT)& --  & 3   & 6   & --  & 9 \\
Long-Horizon (LH/SCALE/SYNTH)    & --  & --  & 4   & 5   & 9 \\
Adversarial (TRAP)               & --  & --  & 7   & 1   & 8 \\
Information Retrieval (IR)       & 1   & 3   & 4   & --  & 8 \\
Policy (POL)                     & 1   & 3   & 4   & --  & 8 \\
Cross-System (CROSS)             & --  & --  & 7   & --  & 7 \\
Specification (SPEC)             & --  & 2   & 5   & --  & 7 \\
Code Review (CR)                 & 1   & 3   & 2   & --  & 6 \\
Distributed (DIST)               & --  & --  & 4   & 2   & 6 \\
Expertise Asymmetry (EA)         & --  & --  & 5   & --  & 5 \\
Negotiation (NEG)                & --  & 1   & 4   & --  & 5 \\
Real-World (GH1--11)             & --  & --  & 11  & --  & 11 \\
\midrule
\textbf{Originally-authored core (153 with ablation)} & \textbf{7} & \textbf{26} & \textbf{104} & \textbf{16} & \textbf{153} \\
Originally-authored extension (no ablation yet) & \multicolumn{4}{c}{\textit{difficulty pending}}                            & 8   \\
Real GitHub bug reports (GH)               & \multicolumn{4}{c}{\textit{medium-difficulty maintainer-reported bugs}}        & 650 \\
Real data-science (RDS)                    & \multicolumn{4}{c}{\textit{hard, parameterized over canonical public datasets}} & 30  \\
Real incident response (RINC)              & \multicolumn{4}{c}{\textit{hard, adapted from public post-mortems}}            & 10  \\
\midrule
\textbf{Total unique templates}            & \multicolumn{4}{c}{}                                                            & \textbf{851} \\
\textit{Seeded evaluation instances}       & \multicolumn{4}{c}{\textit{after multi-seed expansion at the released seeds}}    & \textit{931} \\
\bottomrule
\end{tabular}
\end{table}

\begin{table}[t]
\centering
\caption{\textsc{TeamBench} leaderboard stratified by category tier. Quotas are proportional to category size in the full pool. $^{\ast}$Pool size counts candidate seeded entries eligible for leaderboard selection. Seeded instance counts are reported in Table~\ref{tab:studies}. The per-category breakdown, including which tasks already have complete reference ablation data, is in Appendix~\ref{app:lb90-construction}. Pool size counts candidate seeded entries after category expansion, not unique templates.}
\label{tab:lb90}
\small
\setlength{\tabcolsep}{8pt}
\begin{tabular}{lrrrl}
\toprule
Tier & Categories & Tasks & Pool size$^{\ast}$ & Difficulty mix \\
\midrule
Large ($\geq$ 5 selected)   & 8  & 55 & 854 & Easy 0 / Medium 15 / Hard 29 / Expert 11 \\
Medium (3 to 4 selected)    & 10 & 31 & 72  & Easy 0 / Medium 1 / Hard 25 / Expert 5  \\
Small (1 to 2 selected)     & 3  & 4  & 5   & Easy 0 / Medium 1 / Hard 3 / Expert 0   \\
\bottomrule
\end{tabular}
\end{table}

\begin{table}[h]
\centering
\caption{\textsc{TeamBench} leaderboard per-category quotas. \textit{With abl.}\ is the number of selected tasks that have complete reference 5-condition ablation data. The rest are evaluated for the first time on the leaderboard.}
\label{tab:lb90-detail}
\small
\setlength{\tabcolsep}{6pt}
\begin{tabular}{lcrrl}
\toprule
Refined category & Quota & Pool size & With abl. & Difficulty mix \\
\midrule
Other (Misc, broad GH scrape) & 7  & 422 & 1  & hard:4, medium:3 \\
GitHub Issues (curated)       & 12 & 221 & 12 & medium:12 \\
Real Data Science             & 8  & 90  & 0  & hard:8 \\
Incident Response             & 6  & 26  & 6  & expert:1, hard:5 \\
Security                      & 6  & 32  & 6  & expert:4, hard:2 \\
Software Eng.                 & 6  & 31  & 6  & expert:2, hard:4 \\
Data Engineering              & 5  & 15  & 5  & expert:1, hard:4 \\
Operations                    & 5  & 17  & 5  & expert:3, hard:2 \\
Testing                       & 4  & 12  & 4  & hard:4 \\
Adversarial                   & 3  & 7   & 3  & hard:3 \\
Code Review                   & 3  & 6   & 3  & hard:2, medium:1 \\
Cross-System Integration      & 3  & 5   & 3  & hard:3 \\
Distributed Systems           & 3  & 7   & 3  & expert:3 \\
Information Retrieval         & 3  & 8   & 3  & hard:3 \\
Long-Horizon                  & 3  & 6   & 3  & expert:2, hard:1 \\
Multi-language                & 3  & 6   & 3  & hard:3 \\
Pipeline                      & 3  & 6   & 3  & hard:3 \\
Policy                        & 3  & 9   & 3  & hard:3 \\
Specification                 & 2  & 3   & 2  & hard:1, medium:1 \\
Integration                   & 1  & 1   & 0  & hard:1 \\
Negotiation                   & 1  & 1   & 1  & hard:1 \\
\bottomrule
\end{tabular}
\end{table}

\subsection{Cross-provider role-mixing}
\label{app:rolemix-full}

Table~\ref{tab:rolemix-full} reports all 27 cross-provider configurations on the 25-task stratified subset, three-seed pooled. The top three configurations cluster within 4 points of each other and their Wilson CIs overlap. The cost spread runs from \$0.65 to \$39.58 (all-Anthropic), a $60.9\times$ spread for a 4-point pass-rate spread between the cheapest and most expensive 22.7\%-tier configurations.

\begin{table}[h]
\centering
\caption{All cross-provider configurations on the 25-task subset, sorted by pass rate. Each row pools three seeds at $n{=}75$ runs per configuration. Cost is total USD across all 75 runs. Pass/\$ is pass rate divided by total cost. LLM tool-call turns per session is the mean across $75$ sessions of (Planner steps + Executor steps + Verifier steps + any Executor steps inside Verifier-triggered remediation rounds), where each ``step'' is one assistant message that issues a tool call or returns a verdict. \textbf{Bold} pass rate marks the top configuration, and \textbf{bold} pass/\$ marks the most cost-efficient.}
\label{tab:rolemix-full}
\small
\setlength{\tabcolsep}{3pt}
\begin{tabular}{rlllrrrrr}
\toprule
Rank & Planner & Executor & Verifier & Pass & Partial & Cost \$ & Pass/\$ & Tool-call turns \\
\midrule
1  & \logog~Gemini-3f   & \logoa~Haiku-4.5 & \logoa~Haiku-4.5 & \textbf{26.7\%} & 0.695 & 20.52 & 0.013 & 54.1 \\
2  & \logoo~GPT-5.4m & \logoo~GPT-5.4m & \logoa~Haiku-4.5 & 22.7\% & 0.679 & 10.98 & 0.021 & 30.9 \\
3  & \logog~Gemini-3f   & \logoa~Haiku-4.5 & \logoo~GPT-5.4m & 22.7\% & 0.672 & 11.77 & 0.019 & 49.8 \\
4  & \logoa~Haiku-4.5 & \logoa~Haiku-4.5 & \logoa~Haiku-4.5 & 22.7\% & 0.639 & 39.58 & 0.006 & 54.3 \\
5  & \logoo~GPT-5.4m & \logog~Gemini-3f   & \logog~Gemini-3f   & 21.3\% & 0.612 & 3.73 & 0.057 & 36.2 \\
6  & \logoa~Haiku-4.5 & \logoa~Haiku-4.5 & \logoo~GPT-5.4m & 21.3\% & 0.650 & 29.88 & 0.007 & 45.0 \\
7  & \logoo~GPT-5.4m & \logog~Gemini-3f   & \logoo~GPT-5.4m & 20.0\% & 0.600 & 2.99 & 0.067 & 31.3 \\
8  & \logoa~Haiku-4.5 & \logoo~GPT-5.4m & \logoo~GPT-5.4m & 20.0\% & 0.639 & 9.53 & 0.021 & 37.2 \\
9  & \logoo~GPT-5.4m & \logoo~GPT-5.4m & \logoo~GPT-5.4m & 18.7\% & 0.592 & 2.09 & \textbf{0.089} & 25.2 \\
10 & \logog~Gemini-3f   & \logoo~GPT-5.4m & \logoo~GPT-5.4m & 18.7\% & 0.627 & 2.36 & 0.079 & 38.3 \\
11 & \logoa~Haiku-4.5 & \logog~Gemini-3f   & \logoo~GPT-5.4m & 18.7\% & 0.583 & 12.45 & 0.015 & 46.7 \\
12 & \logoa~Haiku-4.5 & \logog~Gemini-3f   & \logoa~Haiku-4.5 & 18.7\% & 0.618 & 21.62 & 0.009 & 51.7 \\
13 & \logoo~GPT-5.4m & \logoa~Haiku-4.5 & \logoo~GPT-5.4m & 17.3\% & 0.448 & 6.15 & 0.028 & 19.9 \\
14 & \logoo~GPT-5.4m & \logoa~Haiku-4.5 & \logog~Gemini-3f   & 17.3\% & 0.466 & 6.99 & 0.025 & 23.5 \\
15 & \logoa~Haiku-4.5 & \logog~Gemini-3f   & \logog~Gemini-3f   & 17.3\% & 0.605 & 13.34 & 0.013 & 50.9 \\
16 & \logoa~Haiku-4.5 & \logoa~Haiku-4.5 & \logog~Gemini-3f   & 17.3\% & 0.646 & 29.48 & 0.006 & 51.6 \\
17 & \logog~Gemini-3f   & \logog~Gemini-3f   & \logoo~GPT-5.4m & 16.0\% & 0.557 & 3.67 & 0.044 & 45.3 \\
18 & \logoa~Haiku-4.5 & \logoo~GPT-5.4m & \logoa~Haiku-4.5 & 16.0\% & 0.602 & 18.31 & 0.009 & 44.9 \\
19 & \logog~Gemini-3f   & \logoo~GPT-5.4m & \logog~Gemini-3f   & 14.7\% & 0.559 & 3.33 & 0.044 & 45.7 \\
20 & \logog~Gemini-3f   & \logoa~Haiku-4.5 & \logog~Gemini-3f   & 14.7\% & 0.496 & 8.13 & 0.018 & 35.6 \\
21 & \logoa~Haiku-4.5 & \logoo~GPT-5.4m & \logog~Gemini-3f   & 14.7\% & 0.617 & 10.87 & 0.013 & 43.9 \\
22 & \logoo~GPT-5.4m & \logoa~Haiku-4.5 & \logoa~Haiku-4.5 & 14.7\% & 0.437 & 12.41 & 0.012 & 24.2 \\
23 & \logog~Gemini-3f   & \logog~Gemini-3f   & \logog~Gemini-3f   & 13.3\% & 0.570 & 4.98 & 0.027 & 54.6 \\
24 & \logoo~GPT-5.4m & \logog~Gemini-3f   & \logoa~Haiku-4.5 & 13.3\% & 0.399 & 7.98 & 0.017 & 24.5 \\
25 & \logoo~GPT-5.4m & \logoo~GPT-5.4m & \logog~Gemini-3f   & 12.0\% & 0.602 & 2.66 & 0.045 & 29.5 \\
26 & \logog~Gemini-3f   & \logoo~GPT-5.4m & \logoa~Haiku-4.5 & 10.7\% & 0.317 & 7.43 & 0.014 & 32.0 \\
27 & \logog~Gemini-3f   & \logog~Gemini-3f   & \logoa~Haiku-4.5 & 10.7\% & 0.371 & 8.11 & 0.013 & 37.0 \\
\bottomrule
\end{tabular}
\end{table}

\subsection{Leaderboard detail}
\label{app:lb90-leaderboard}

Table~\ref{tab:lb90-leaderboard} reports the full per-condition pass-rate matrix that backs Figure~\ref{fig:leaderboard}. The leaderboard panel distinguishes models with Full coverage at $n{\geq}30$ from those below threshold (rows below the divider). Aggregation deduplicates by (model, condition, task) across canonical and resumed-checkpoint result files, with later writes winning.

\begin{table}[h]
\centering
\caption{\textsc{TeamBench} leaderboard per-condition pass rates (\%). Models are sorted by $\max($Solo, Full$)$ descending. \textbf{Bold} marks the highest cell across the row.}
\label{tab:lb90-leaderboard}
\small
\setlength{\tabcolsep}{6pt}
\begin{tabular}{lrrrrr}
\toprule
Model & Solo & Restricted & No Plan & No Eval & Full \\
\midrule
\logoa~Claude Opus 4.7         & 35.6 & 33.3 & 35.6 & 33.3 & \textbf{37.8} \\
\logoo~GPT-5.4 Mini            & \textbf{33.3} & 23.3 & 25.6 & 24.4 & 28.9 \\
\logoa~Claude Haiku 4.5        & 12.2 & \textbf{31.1} & 18.9 &  1.1 & 28.9 \\
\logog~Gemini-3.1 Pro          & 27.8 & 22.2 & 16.7 & 25.6 & \textbf{28.9} \\
\logoa~Claude Sonnet 4.6       &  7.8 & \textbf{27.8} & 10.0 &  6.7 & \textbf{27.8} \\
\logoo~GPT-5.4                 & 12.2 & \textbf{35.6} & 23.3 & 34.4 & 27.8 \\
\logog~Gemma 4 31B             & \textbf{27.8} & 25.6 & 24.4 & 20.0 & 22.2 \\
\logog~Gemini-3 Flash          & 13.3 & 18.9 & 14.4 & \textbf{27.8} & 25.6 \\
\logog~Gemini-3.1 Flash Lite   &  5.6 & \textbf{21.1} &  8.9 & 17.8 & 17.8 \\
\logoo~gpt-oss-20b             & \textbf{17.8} & 17.8 & 12.2 &  7.8 &  2.2 \\
\logoq~Qwen 3 14B              & \textbf{5.6} &  2.2 &  2.2 &  1.1 &  2.2 \\
\logoq~Qwen 3 32B              & \textbf{5.6} &  3.3 &  0.0 &  5.6 &  1.1 \\
\logoq~Qwen 3 8B               &  2.2 & \textbf{5.6} &  1.1 &  3.3 &  0.0 \\
\bottomrule
\end{tabular}
\end{table}

\footnotetext{The default grader marks a run as a failure whenever any check fails, including the attestation check that only verifies the agent wrote a valid attestation JSON at the end of the run. In Solo mode the attestation is the agent's self-confirmation and does not measure task quality. Several frontier models systematically forget the attestation file even when every structural check passes. The promotion rule used here counts a run as a pass if all failures are attestation-related and no other failure mode is present, while keeping the original verdict alongside the promoted verdict so the operation is fully reversible. The released aggregate records both verdicts so the headline numbers can be reproduced under either rule.}

The Qwen 3 family scores near zero across all conditions and the gpt-oss-20b Full collapses to $2.2\%$ despite a higher Solo of $17.8\%$. The dominant cause is malformed tool calls and context overflow rather than raw capability (Appendix~\ref{app:open-source}).

\section{Verified Subset and Reproducibility}
\label{app:verified-audit-repro}

% \subsection{Verified Subset}
% \label{sec:verified}

We audit whether the graders support the reported outcomes. The audit has four components. A canonical solution check, following the idea of SWE-Bench Verified, records whether a known correct solution passes the grader. Mutation testing applies AST-level mutations to a passing workspace and asks whether the grader catches them. Cross-model discrimination flags tasks where every leaderboard model produces the same outcome. A grader plausibility audit asks three independent LLM-judges to label 285 stratified runs as PASS or FAIL. We run this audit in two variants. With the deterministic verdict shown to judges, Fleiss's $\kappa = 0.74$. With the verdict hidden (a leakage-free re-run on the same $285$ tuples, Appendix~\ref{app:verified-detail}), Fleiss's $\kappa = 0.07$. The $0.74$ value is therefore largely anchored by the verdict, and we report \emph{neither} variant as independent grader validation. The leakage-free run surfaces a systematic disagreement: Gemini-3 Flash returns PASS on $53\%$ of cases against the grader's $14\%$, mirroring its over-acceptance as a Verifier (\S\ref{sec:ref-results}). \textbf{TeamBench-Verified} requires the canonical-solution and discrimination checks, and applies mutation testing where source-level mutation is applicable. $57$ tasks ($63.3\%$ of the leaderboard) clear these thresholds. ``Verified'' means \emph{audited} on the pillars that apply rather than expert-verified per task. On the role mixing pool restricted to the Verified subset, the Verifier false accept rate is 38.7\%. The same failure pattern therefore appears on the audited subset (Appendix~\ref{app:verified-detail}).

\subsection{Per-pillar detail}
\label{app:verified-detail}

Table~\ref{tab:verified-pillars} gives per-pillar eligibility, passing denominators, and the rule applied to non-eligible tasks. In this submission, “Verified” means that a task passes the applicable audit checks. It does not mean that every task has been expert adjudicated.

\begin{table}[h]
\centering
\caption{Per-pillar audit denominators for the leaderboard subset.}
\label{tab:verified-pillars}
\small
\setlength{\tabcolsep}{4pt}
\begin{tabularx}{\linewidth}{l c X X}
\toprule
Pillar & Eligible / passing & Rule for non-eligible & Notes \\
\midrule
Canonical solution     & $90$ / $58$ & required & $58$ pass the LLM-run evidence path \\
Mutation testing       & $9$ / $7$   & exempt where AST mutation is inapplicable & $1$ task at kill-rate $0.42$ excluded \\
Cross-model discrim.   & $90$ / $58$ & required & spread $\geq 0.10$ across the leaderboard panel \\
Grader-plausibility    & $285$ / $-$ & sampled audit only, not eligibility-gating & Fleiss's $\kappa=0.74$ with verdict, $0.07$ without \\
\bottomrule
\end{tabularx}
\end{table}

\paragraph{Canonical-solution check.} A task is verified if some known-correct solution exists that the deterministic grader actually accepts. We accept three sources of proof, in order: the seed-0 workspace as-is, any historical agent run whose post-run workspace passed the grader, or the upstream PR diff applied to a GH-sourced task. $58$ of $90$ leaderboard tasks pass via the LLM-run evidence path. We label this an audited rather than expert-verified evidence path because canonical solutions are not always human-authored.

\paragraph{Mutation-killing grader.} For each task with a known-passing workspace, we apply small AST-level mutations (operator swaps, return-value flips, body deletions) and check whether the grader catches them. The threshold is mutation kill rate $\geq 0.5$. Of the leaderboard tasks with a known-passing workspace ($n=9$), $7$ exceed $0.5$ and $4$ exceed $0.7$. \texttt{O6\_perf\_tuning} is exempt by design, since its grader scores a deliverable artifact via a simulator rather than running workspace source code. Tasks without a known-passing workspace are not eligible for source-level mutation testing.

\paragraph{Cross-model discrimination.} Tasks where every leaderboard model produces the same outcome carry no comparative signal. We require a per-task pass-rate spread across the leaderboard panel of $\geq 0.10$. $58$ of $90$ leaderboard tasks meet this bar.

\paragraph{Grader-plausibility audit (with-verdict and leakage-free).} We run the same $285$ stratified judgments under two prompt variants. \emph{With-verdict}: judges see the spec, a workspace summary, the verifier attestation, and the deterministic-grader verdict. \emph{No-verdict}: the same $285$ tuples with the deterministic verdict removed from the prompt, so each judge forms its own opinion from spec and artifacts alone. Table~\ref{tab:rater-leakage} reports both. Cross-judge agreement falls from substantial ($\kappa = 0.74$) under the with-verdict prompt to slight ($\kappa = 0.07$) under the leakage-free prompt, indicating that the original $0.74$ figure was anchored by the verdict. We therefore use \emph{neither} audit as independent grader validation. Two patterns survive both variants and align with the main paper. First, Haiku 4.5 and GPT-5.4 Mini track the deterministic grader's PASS rate ($14\%$ overall) reasonably well even without the verdict (PASS rates of $11\%$ and $10\%$ respectively, agreement $82\%$ and $83\%$). Second, Gemini-3 Flash returns PASS on $53\%$ of cases when the verdict is hidden, mirroring its $77\%$ false-accept rate as a Verifier (\S\ref{sec:ref-results}). The honest reading is that the audit is informative about how each model judges pass/fail in isolation and consistent with the agent-side Verifier-fails finding, not that it independently validates the deterministic grader.

\begin{table}[h]
\centering
\caption{LLM grader-plausibility audit on $285$ stratified runs, with the deterministic verdict shown to judges (left) and hidden from judges (right). Substantial-looking agreement under the original protocol is largely anchored by the verdict, while the leakage-free protocol reveals systematic disagreement, particularly Gemini-3 Flash's over-acceptance.}
\label{tab:rater-leakage}
\small
\setlength{\tabcolsep}{6pt}
\begin{tabular}{lcc}
\toprule
Quantity & With verdict (original) & No verdict (leakage-free) \\
\midrule
Three-way Fleiss's $\kappa$           & $0.74$  & $0.07$ \\
Binary Krippendorff's $\alpha$        & $0.74$  & $0.07$ \\
Pairwise Cohen's $\kappa$, Haiku $\leftrightarrow$ Gemini & $0.66$ & $0.18$ \\
Pairwise Cohen's $\kappa$, Haiku $\leftrightarrow$ GPT   & $0.91$ & $0.27$ \\
Pairwise Cohen's $\kappa$, Gemini $\leftrightarrow$ GPT  & $0.70$ & $0.10$ \\
Haiku 4.5 PASS rate                   & $14.0\%$ & $10.9\%$ \\
Gemini-3 Flash PASS rate              & $14.7\%$ & $53.3\%$ \\
GPT-5.4 Mini PASS rate                & $13.7\%$ & $9.5\%$ \\
Deterministic grader PASS rate        & $14.4\%$ & $14.4\%$ \\
\bottomrule
\end{tabular}
\end{table}

\paragraph{Triage of non-Verified tasks.} Of the non-Verified tasks, $11$ are \emph{near-miss-very-close} (best historical partial $\geq 0.9$ but no run passes), $11$ are \emph{near-miss} (partial $0.7$ to $0.9$), $10$ are \emph{solvable-in-principle} (partial $0.4$ to $0.7$), and one (\texttt{TRAP1\_spec\_conflict}) sits at mutation kill rate $0.42$, just below threshold. The per-task ledger is part of the public release.

\paragraph{Verifier false-accept robustness on the audited subset.}

% \begin{table}[h]
% \centering
% \caption{Verifier false-accept rate on the role-mixing pool restricted to the \textsc{TeamBench-Verified} subset. Computed by walking each role-mixing run's submission attestation and pairing it with the deterministic grader's verdict at the run level. The denominator differs slightly from the $1{,}083$-run aggregated figure in \S\ref{sec:ref-results} ($49.4\%$) because the run-level walk uses a different deduplication pass. Both numbers reflect the same underlying claim that LLM Verifiers approve roughly half of grader-failing runs. The Verifier-fails finding survives the audited-subset restriction. The Q1/Q5 quintile contrast (reference ablation) and the prompt-only-vs-enforced contrast ($25$-task subset) use task pools that only partially intersect the $57$-task Verified subset and are reported on their original pools.}
% \label{tab:verified-robustness}
% \small
% \setlength{\tabcolsep}{6pt}
% \begin{tabular}{lccc}
% \toprule
% Headline claim & Pool / $n$ & Full pool & TeamBench-Verified \\
% \midrule
% Verifier false-accept rate (role-mixing) & grader-fail $798$ vs.\ $530$ & $51.7\%$ & $38.7\%$ \\
% \bottomrule
% \end{tabular}
% \end{table}

% \section{Comparison Against Prior Benchmarks}
% \label{app:comparison}

\begin{table}[h]
\centering
\caption{Comparison with related benchmarks and benchmark ecosystems across eight design axes. SA = single-agent, MA = multi-agent. \gcmark\,= supported, \rxmark\,= not supported, \partialmark\,= partial or planned. The first four axes describe what the benchmark measures. The last four describe whether the benchmark is versioned, holds out evaluation seeds, releases run logs, and is scheduled for periodic refresh.}
\label{tab:benchmark-comparison}
\small
\resizebox{\columnwidth}{!}{%
\begin{tabular}{clcccccccc}
\toprule
& \textbf{Benchmark} & \textbf{Struct.\ enf.} & \textbf{Role abl.} & \textbf{Cross-prov.} & \textbf{Contam.\ res.} & \textbf{Vers./live} & \textbf{Verified} & \textbf{Human bsl.} & \textbf{Pub.\ traces} \\
\midrule
\multirow{7}{*}{\rotatebox[origin=c]{90}{\small SA}}
  & SWE-Bench~\citep{jimenez2024swebench}      & --      & \rxmark & \gcmark & \rxmark & \gcmark      & \gcmark      & \rxmark      & \partialmark \\
  & Terminal-Bench~\citep{terminalbench2026}    & --      & \rxmark & \gcmark & \gcmark & \gcmark      & \rxmark      & \rxmark      & \partialmark \\
  & LiveCodeBench~\citep{jain2024livecodebench}    & --      & \rxmark & \gcmark & \gcmark & \gcmark      & \rxmark      & \rxmark      & \rxmark      \\
  & GAIA~\citep{mialon2023gaia}             & --      & \rxmark & \gcmark & \rxmark & \rxmark      & \rxmark      & \gcmark      & \rxmark      \\
  & MLE-Bench~\citep{chan2024mlebench}        & --      & \rxmark & \gcmark & \rxmark & \rxmark      & \rxmark      & \gcmark      & \partialmark \\
  & BrowseComp~\citep{browsecomp2025}       & --      & \rxmark & \gcmark & \rxmark & \rxmark      & \rxmark      & \gcmark      & \rxmark      \\
  & AgentBench~\citep{liu2024agentbench}    & --      & \rxmark & \gcmark & \rxmark & \rxmark      & \rxmark      & \rxmark      & \rxmark      \\
\midrule
\multirow{4}{*}{\rotatebox[origin=c]{90}{\small MA}}
  & MultiAgentBench~\citep{zhu2025multiagentbench}  & \rxmark & \rxmark      & \gcmark      & \rxmark & \rxmark      & \rxmark      & \rxmark      & \rxmark \\
  & DevBench~\citep{li2024devbench}         & \rxmark & \rxmark      & \partialmark & \rxmark & \rxmark      & \rxmark      & \rxmark      & \rxmark \\
  & ChatDev~\citep{qian2024chatdev}         & \rxmark & \rxmark      & \rxmark      & \rxmark & \rxmark      & \rxmark      & \rxmark      & \rxmark \\
  & GPTSwarm~\citep{zhuge2024gptswarm}       & \rxmark & \rxmark      & \rxmark      & \rxmark & \rxmark      & \rxmark      & \rxmark      & \rxmark \\
\midrule
  & \textbf{\textsc{TeamBench} (Ours)} & \gcmark & \gcmark & \gcmark & \gcmark & \partialmark & \gcmark & \partialmark & \gcmark \\
\bottomrule
\end{tabular}%
}
\end{table}

\subsection{Reproducibility}
\label{app:reproducibility}

\paragraph{Code and data availability.}
The benchmark, all generators, the evaluation harness, the role-mixing protocol, and the 5-condition ablation runner are released under a permissive open-source license at \url{https://teambench.github.io/}. The leaderboard task selection JSON, the per-condition reference-ablation logs, the $2{,}025$ deduplicated role-mixing per-task records, and the cost ledger are included in the release. The Docker images for the Planner, Executor, and Verifier roles are built with deterministic dependencies pinned by hash.

\paragraph{Determinism and replication.}
All graders are deterministic shell scripts. All API calls use temperature 0. Task generators take a seed and produce byte-identical workspaces across machines. Replication of the reference ablation requires 1{,}165 task runs under gemini-3-flash-preview. Replication of the cross-provider role-mixing study requires 2{,}025 task runs across three commercial providers (27 configurations $\times$ 25 tasks $\times$ 3 seeds), with measured total spend of \$326.04.

\paragraph{Held-out seeds and contamination.}
Seeds 0 through 2 are released with the public benchmark. Seeds 5 and above are reserved for the hidden leaderboard. Across $50$ tasks, $72\%$ of held-out seed workspaces differ on every file from their development counterparts and the mean Jaccard similarity is $0.71 \pm 0.23$. The leaderboard runner re-rolls task seeds at submission time using a leaderboard-only RNG state. Parameterized seeds reduce exact value memorization but do not eliminate semantic contamination, so for studies whose primary claim is memorization resistance we recommend evaluating on the held-out leaderboard seeds rather than on public seeds.

\paragraph{Responsible use.}
The adversarial-trap and security-vulnerability tasks contain plausible-but-incorrect security patterns by design. They are synthetic evaluation cases and are not designed to target real systems. The cryptographic tasks use intentional weaknesses (nonce reuse, low PBKDF2 iterations, truncated authentication tags) for evaluation. These patterns must not be deployed. We include a per-task usage note in the task metadata that flags adversarial content, and we recommend that hosted leaderboard submissions run in network-isolated containers.

\paragraph{Dataset card and metadata.}
A dataset card following \citet{gebru2021datasheets} accompanies the public release. It documents the collection process, the annotation procedure for difficulty and category labels, intended uses, known biases (English-language tasks only, Python-heavy), and a contact channel for issues. The dataset card is versioned with the release tags so that any update to a task triggers a card revision. The dataset is hosted on Hugging Face at \url{https://huggingface.co/datasets/ybkim95/teambench} and ships with Croissant~$1.1$ machine-readable metadata, including Responsible AI fields. Both the dataset content and the code release are distributed under the MIT license.

\paragraph{Benchmark governance plan.}
Table~\ref{tab:governance} summarizes the governance features that ship with the initial release and the items scheduled for the first follow-up.

\begin{table}[h]
\centering
\caption{Benchmark governance plan. \gcmark\ ships with the initial release. \partialmark\ is scheduled for the first follow-up release.}
\label{tab:governance}
\small
\setlength{\tabcolsep}{6pt}
\begin{tabular}{p{4.6cm}cp{6.8cm}}
\toprule
Feature & Status & Notes \\
\midrule
Public code, generators, harness         & \gcmark      & MIT license. \\
Public per-task run records              & \gcmark      & 1{,}165 reference-ablation runs and 2{,}025 role-mixing runs (27 configs $\times$ 25 tasks $\times$ 3 seeds). \\
Hidden held-out seeds                    & \gcmark      & Seeds 5 and above. \\
Versioned releases (semver)              & \gcmark      & v1.0 ships with this paper. \\
Issue tracker for problematic tasks      & \gcmark      & GitHub issues, with per-task usage notes. \\
Submission protocol                      & \gcmark      & JSON contract documented in the release. \\
Canary string in public docs             & \partialmark & To discourage training-data ingestion. \\
\textsc{TeamBench-Verified-Human} subset & \partialmark & Future expert-adjudicated subset, distinct from the leaderboard. \\
\textsc{TeamBench-Live} monthly refresh  & \partialmark & Newly-generated GitHub issues. \\
Compare-by-cost leaderboard view         & \partialmark & Already collected. \\
Public trace download                    & \partialmark & Per-run agent transcripts. \\
\bottomrule
\end{tabular}
\end{table}

The three planned-status rows of Table~\ref{tab:governance} are the prioritized follow-up items: an expert-adjudicated \textsc{TeamBench-Verified-Human} subset, a canary string for contamination audits, and a public per-run trace download.

\subsection{Adherence to the Agentic Benchmark Checklist}
\label{app:abc}

We map each criterion of the Agentic Benchmark Checklist (ABC) of \citet{zhu2025abc} to the corresponding TeamBench design choice (Table~\ref{tab:abc}). Items that do not apply to a deterministic shell-script grader (fuzzing) are marked N/A.

\begin{table}[h]
\centering
\small
\setlength{\tabcolsep}{4pt}
\caption{TeamBench against the Agentic Benchmark Checklist. T = task validity, O = outcome validity, R = reporting.}
\label{tab:abc}
\begin{tabular}{p{0.7cm}p{4.6cm}p{8.5cm}}
\toprule
ID & Criterion & TeamBench design choice \\
\midrule
\multicolumn{3}{l}{\textit{Task validity}} \\
T.1 & Specify tool / package versions & Docker images pin all dependencies by hash; per-task setup script lists required toolchain \\
T.2 & Manage service availability and rate limits & Harness retries on 429 / 503 with key rotation; infrastructure failures count as task failures under the $n=90$ convention \\
T.3 & Detect API interruptions and terminate evaluation & Per-run logs capture HTTP status; failed runs are recorded with error code \\
T.4 & Clean up legacy state & Each run starts in a fresh container with the seed-0 workspace \\
T.5 & Isolate agents from ground truth & Grader runs in a separate container; expected outputs are written to a grader-only directory the agent cannot read \\
T.6 & Reproducible environment & Hash-pinned Docker images; deterministic shell-script graders \\
T.7 & Verify ground-truth annotation & Canonical-solution audit verifies a passing workspace exists for $58$ of $90$ LB90 tasks \\
T.8 & Verify task solvability & Same canonical-solution check; the 5-condition ablation pool achieves non-zero per-condition pass on every task \\
T.9 & Provide oracle solver & Canonical-via-LLM-run evidence path serves as the solver demonstration \\
T.10 & Inspect outliers in pilot & Pilot evidence ledger; outlier sessions documented \\
\midrule
\multicolumn{3}{l}{\textit{Outcome validity}} \\
O.a.1 & Semantically equivalent expressions & Graders use exact-string or deterministic equivalence on JSON / structured fields \\
O.a.2 & Handle redundant words & Graders normalize whitespace; output formats are documented in the spec \\
O.b.1 & Negation modifiers & Policy and spec tasks include negation tests in the grader \\
O.b.2 & Prevent listing-all-answers success & Multi-check graders fail most checks if every option is listed \\
O.b.3 & Avoid guessing success & Multiple grader checks per task; partial-score grader rewards specific outputs \\
O.c.1 & Pilot LLM-judge accuracy & 285-tuple LLM-judge audit (App D.1) \\
O.d.1 & Manually verify test cases & Test cases are maintainer-reported (GitHub-curated) or author-designed for originally-authored tasks \\
O.d.2 & Coverage as quality metric & Per-task graders use multiple specific checks (median 10 across the originally-authored set) \\
O.e.1--3 & Fuzzing comprehensiveness & N/A. Tasks are spec-driven, not fuzzed \\
O.f.1 & Cover all branches of user workflows & Multiple checks per task; full pass requires every check to pass \\
O.f.2 & Eliminate non-determinism & Deterministic graders; temperature 0; fixed seeds \\
O.g.1 & Ground-truth all outcomes & Specs enumerate accepted outputs; partial-score grader awards partial progress \\
O.g.2 & Relevant + irrelevant states & Workspaces include distractor files (Adversarial / Trap categories) \\
O.g.3 & Sufficiently complex state space & Workspaces span 0 to 40 files; multi-file edits required \\
O.h.1 & Explicit format assumptions & Specs document required output formats \\
O.h.2 & Avoid guessing success & Same as O.b.3 \\
O.i.1 & Metrics correlate with reasoning & Partial-score checks tied to specific behaviors documented in spec \\
\midrule
\multicolumn{3}{l}{\textit{Benchmark reporting}} \\
R.1 & Open-source dataset and harness & Public release on Hugging Face and GitHub \\
R.2 & Open evaluation harness & Same release \\
R.3 & Contamination prevention & Held-out seeds $\geq 5$, parameterized generators, Croissant $1.1$ metadata \\
R.4 & Plan for consistent updates & Versioned releases; governance plan in Table~\ref{tab:governance} \\
R.5 & Specify capabilities & Planner / Executor / Verifier capabilities in Table~\ref{tab:conditions} \\
R.6 & Articulate construct validity & §3 ties each ablation to the role-marginal it measures \\
R.7 & Document mitigation & §4 and §5 lists scope and limitations \\
R.8 & Quantitative limitation impact & TeamBench-Verified rate $38.7\%$ vs full-pool $49.4\%$ quantifies the grader-completeness impact \\
R.9 & Quantitative impact & Same as R.8 \\
R.10 & Statistical significance & Wilson 95\% CIs, paired bootstrap (10{,}000 iterations), McNemar with Holm-Bonferroni \\
R.11 & Interpretation guidelines & §3.2 and §3.4 spell out interpretation \\
R.12 & Baseline comparisons & Solo / Restricted / No-Plan / No-Evaluate / Full Team \\
R.13 & Trivial agent baselines & Restricted is the trivial baseline \\
\bottomrule
\end{tabular}
\end{table}

\section{Additional Results}
\label{app:additional-analyses}

\subsection{Per-category teamwork effect}
\label{app:percat}

\begin{figure}[h]
    \centering
    \includegraphics[width=0.85\linewidth]{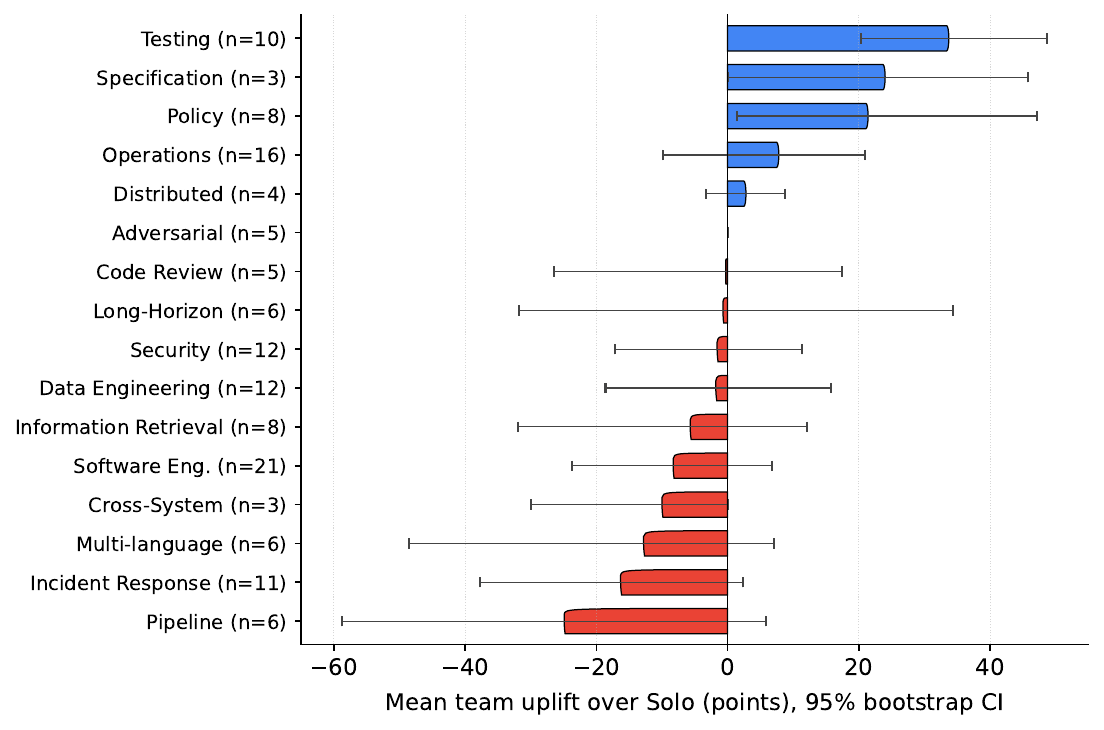}
    \caption{Per-category mean team uplift on the reference 5-condition ablation. Uplift is full-team partial score minus Solo partial score. Blue indicates positive uplift, red indicates negative. Bars include 95\% bootstrap CIs (10{,}000 iterations). The figure includes only categories with three or more tasks for stable intervals. Per-category counts use the canonical task-to-category mapping in the released dataset.}
    \label{fig:percat}
\end{figure}

The categories with the largest positive bootstrap means are Testing, Specification, and Policy. The categories with the largest negative means are Pipeline, Incident Response, and Multi-language. Categories with positive uplift share a structural property in which the specification carries decision rules that the Executor cannot independently derive from the workspace alone. The wide CIs on small-$n$ categories (for example Specification at $n{=}3$ and Cross-System at $n{=}3$) overlap zero, so those category-level effects should be read as suggestive rather than conclusive.

\subsection{Verifier confusion matrix}
\label{app:evaluator-confusion}

We extract the Verifier verdict from each run's attestation file and the deterministic grader result from the grader's score record for every role-mixing run that produced both. Table~\ref{tab:evaluator-confusion} reports the per-Verifier-provider breakdown on the complete pool of 1{,}083 attestation-bearing runs across all 27 configurations and three seeds (2{,}025 unique role-mixing runs minus 942 that lacked an attestation due to harness errors, run timeouts, or pending verifier turns). The pooled false-accept rate of 49.4\% (Wilson 95\% CI $[45.9, 52.9]$) matches the headline reported in the Discussion (Section~\ref{sec:discussion}).

\begin{table}[h]
\centering
\caption{Verifier confusion matrix on the 1{,}083 role-mixing runs with an attestation and a grader score, broken down by the Verifier's provider family. \textit{Actual pass} is the deterministic grader result. \textit{Verdict} is the Verifier's verdict (pass or fail). The Verifier accepts incomplete work between 36\% and 77\% of the time across providers and almost never rejects correct work.}
\label{tab:evaluator-confusion}
\small
\begin{tabular}{lrrrr}
\toprule
Verifier provider & TP & FP & FN & TN \\
\midrule
\logoa~Haiku-4.5 (n=331) & 98  & 140 & 0  & 93  \\
\logog~Gem-3f (n=199)    & 70  & 87  & 16 & 26  \\
\logoo~GPT-5.4m (n=553)  & 117 & 157 & 4  & 275 \\
\midrule
\textbf{Pooled (n=1{,}083)}  & \textbf{285} & \textbf{384} & \textbf{20} & \textbf{394} \\
\bottomrule
\end{tabular}

\vspace{4pt}
\begin{tabular}{lrrr}
\toprule
Metric & Pooled & GPT (best) & Gemini (worst) \\
\midrule
Accuracy           & 62.7\% & 70.9\% & 48.2\% \\
False-reject rate  & 6.6\%  & 3.3\%  & 18.6\% \\
False-accept rate  & 49.4\% & 36.3\% & 77.0\% \\
\bottomrule
\end{tabular}
\end{table}

The pattern across the three providers is consistent. The Verifier rarely rejects valid work, with the pooled false-reject rate at $6.6\%$ and the per-provider rates between $0\%$ and $19\%$. The Verifier accepts incomplete work between $36\%$ and $77\%$ of the time. GPT-5.4 Mini is the most discriminative Verifier in this slot. Gemini-3 Flash is the least discriminative, accepting $77\%$ of grader-failing runs and producing the worst overall accuracy at $48\%$. The headline implication is that current Verifiers in this single-pass file-based protocol return pass by default rather than functioning as a quality gate.

\paragraph{Missingness sensitivity.} The pool of $1{,}083$ attestation-bearing runs excludes $942$ runs of the $2{,}025$-cell role-mixing grid where the agent did not produce a valid attestation. Treating those $942$ missing attestations as failures rather than excluding them (Table~\ref{tab:evaluator-missingness}) lowers the effective false-accept rate but does not change the qualitative finding that Verifiers do not function as a quality gate, because the missing-attestation runs are themselves verifier-side or system-side failures.

\begin{table}[h]
\centering
\caption{Verifier-failure rate under different treatments of the $942$ runs that lacked a valid attestation.}
\label{tab:evaluator-missingness}
\small
\setlength{\tabcolsep}{6pt}
\begin{tabular}{lll}
\toprule
Treatment of missing attestations & Effective rate & Interpretation \\
\midrule
Exclude (attestation-bearing only) & $49.4\%$ false-accept among grader-failing & verdict-quality bound \\
Treat as fail verdict              & $22.3\%$ false-accept among grader-failing & conservative bound \\
Count as verifier-side failure     & $66.5\%$ verifier failure overall          & reliability bound \\
\bottomrule
\end{tabular}
\end{table}

\subsection{Mapping MAST failure modes to TeamBench detectors}
\label{app:mast-mapping}

Table~\ref{tab:mast-mapping} maps the 14 failure modes of \citet{cemri2025whyfail} to the role-violation detectors and grader signals used in TeamBench. Modes that depend on multi-turn dialogue do not apply to the single-pass file-based protocol and are marked N/A.

\begin{table}[h]
\centering
\small
\setlength{\tabcolsep}{4pt}
\caption{MAST failure modes mapped to TeamBench detectors. Quoted definitions are from \citet{cemri2025whyfail}.}
\label{tab:mast-mapping}
\begin{tabular}{p{1.0cm}p{4.5cm}p{8.0cm}}
\toprule
MAST & Failure mode & TeamBench detector / signal \\
\midrule
\multicolumn{3}{l}{\textit{System design issues}} \\
FM-1.1 & Disobey task specification & verifier-modifies-code event; deterministic-grader fail \\
FM-1.2 & Disobey role specification & verifier-modifies-code, planner-writes-code, executor-self-approves \\
FM-1.3 & Step repetition & per-run turn count; not a primary detector \\
FM-1.4 & Loss of conversation history & N/A (single-pass file-based protocol) \\
FM-1.5 & Unaware of termination conditions & missing-attestation rate \\
\midrule
\multicolumn{3}{l}{\textit{Inter-agent misalignment}} \\
FM-2.1 & Conversation reset & N/A (no multi-turn dialogue) \\
FM-2.2 & Fail to ask for clarification & N/A (single-turn) \\
FM-2.3 & Task derailment & Optimistic verdict failures (one-line approvals) \\
FM-2.4 & Information withholding & Planner-to-Executor relay fidelity 0.21 (§\ref{sec:capability}) \\
FM-2.5 & Ignored other agent's input & verifier-modifies-code (Verifier ignores Executor evidence) \\
FM-2.6 & Reasoning-action mismatch & Echo verdicts (Verifier reasons completeness without checking) \\
\midrule
\multicolumn{3}{l}{\textit{Task verification}} \\
FM-3.1 & Premature termination & Optimistic-verdict subtype: one-line approvals \\
FM-3.2 & No or incomplete verification & 49.4\% Verifier false-accept rate (§\ref{sec:ref-results}) \\
FM-3.3 & Incorrect verification & False-reject rate 6.6\% pooled; Gemini-3 Flash 18.6\% (App F) \\
\bottomrule
\end{tabular}
\end{table}

\subsection{Open-source failures}
\label{app:open-source}

The smallest open-source models on the \textsc{TeamBench} leaderboard score near zero across all conditions. Three failure modes account for nearly all the failures.

\begin{itemize}[nosep]
\item \textbf{Malformed tool calls.} The model emits text that resembles a tool call (for example \texttt{<tool\_call>...}) but does not conform to the expected JSON schema. The harness records a no-op turn.
\item \textbf{Output-budget exhaustion.} Some reasoning models emit long reasoning blocks that exhaust the harness's per-turn output budget (8,192 tokens) before producing any tool calls.
\item \textbf{Missing attestation.} The agent produces output but fails to write a valid \texttt{attestation.json}. The grader reports an automatic failure regardless of workspace changes.
\end{itemize}

Tool-use reliability is a binding constraint at the small open-weight tier rather than raw language-modeling capability. Malformed tool calls and context overflow are counted as task failures rather than errors excluded from the denominator (Table~\ref{tab:lb90-leaderboard}). We have implemented lenient parsing for several common malformed patterns. The aggregate impact is that every Qwen 3 row and the gpt-oss-20b Full cell collapse to under $3\%$, an order of magnitude below the smallest fully-functional commercial models. At this tier, \textsc{TeamBench} measures tool-call reliability more than coordination, which we report as a benchmark limitation.

\subsection{Expertise asymmetry}
\label{app:ea-results}

\begin{table}[h]
\centering
\caption{Expertise-Asymmetry results pooled across three Gemini models ($N=200$ total runs). Analysis Value = Full $-$ No-Analysis. All five tasks show positive analysis value. EA5 (dependency audit) benefits most.}
\label{tab:ea-detail}
\small
\begin{tabular}{llcccc}
\toprule
Task & Planner Tool & Full & No-Anal. & Solo & Anal.\ Value \\
\midrule
EA1: Security Scan & \texttt{bandit}      & 50.6\% & 30.7\% & 14.5\% & $+19.9$ \\
EA2: Coverage Gap  & \texttt{pytest-cov}  & 78.8\% & 71.2\% & 66.2\% & $+7.7$  \\
EA3: Type Safety   & \texttt{mypy}        & 58.8\% & 50.4\% & 48.2\% & $+8.5$  \\
EA4: Code Quality  & \texttt{ruff/pylint} & 50.0\% & 34.6\% & 30.8\% & $+15.4$ \\
EA5: Dep.\ Audit   & \texttt{pip-audit}   & 64.8\% & 37.8\% & 23.7\% & $+27.0$ \\
\midrule
\textit{Pooled mean} & & 60.6\% & 44.9\% & 36.7\% & $+15.7$ \\
\bottomrule
\end{tabular}
\end{table}

The pooled TNI across three models is 0.265 (bootstrap 95 percent CI $[0.170, 0.356]$), entirely above zero. On this five-task exploratory subset, tool-augmented teamwork provides statistically significant benefit. A TNI exceeding 1.0 is achieved by Gemini-3 Flash Preview (1.25), which means tool-specialized teams can exceed the full-access Solo baseline on this subset. The implication is that multi-agent systems can be designed around complementary capabilities, not just complementary information.

\subsection{Equalizer mechanism}
\label{app:equalizer-mechanism}

We test three candidate mechanisms for the conditional team-uplift pattern reported in Section~\ref{sec:capability}, on a $472$-pair cross-model panel.

\paragraph{H1: Specification relay.}
If team benefit stems from the Planner relaying specification knowledge, tasks with larger necessity gaps (Solo minus Restricted) should show greater team uplift. The opposite holds: necessity gap is negatively correlated with team uplift ($r = -0.446$, $p < 10^{-24}$, $n = 472$). Spec relay alone is not sufficient to explain the pattern.

\paragraph{H2: Step-limit exhaustion.}
If single agents exhaust their step budget on hard tasks while teams distribute the budget across roles, Solo elapsed time should predict failure. Solo elapsed time shows zero correlation with Solo score ($r = 0.00$, $p = 0.99$), and weak models fail equally fast across difficulties (hard / easy elapsed ratio $= 0.93$). Not supported.

\paragraph{H3: Implicit chain-of-thought.}
If the Planner's structured output functions as implicit chain-of-thought for the Executor, the No-Plan condition should underperform Restricted on hard tasks. Empirically, No-Plan and Restricted are statistically indistinguishable on hard tasks (mean difference $-0.006$, $p = 0.66$, $n = 202$).

\paragraph{Capability-conditional uplift.}
A quadratic model of team uplift versus Solo score ($R^2 = 0.60$) outperforms a linear model ($R^2 = 0.40$) on the $472$-pair panel, and the parametric vertex sits at Solo $0.54$ with a fitted peak of $7.6$ points ($p = 0.002$). The equal-size quintile analysis on the originally-authored reference ablation reported in Section~\ref{sec:capability} places the peak at Q1 (hardest). The two statistics answer different questions on different pools and we report both.

\subsection{Enforcement sensitivity}
\label{app:enforcement-sensitivity}

We re-run the pre-specified outcome contrasts (T2, T3) on the $450$ planned cells after substituting the $50$ excluded runs with deterministic failure outcomes. The substitution barely changes prompt-only ($150$ of $150$ already observed) and enforced (only $2$ of $150$ missing), but shifts enforced-shared-history substantially because the Gemini-3 Flash quota incident affected $48$ of $150$ cells. Pass rate moves from $48.0\%$ (observed) to $32.7\%$ (with missing as fail). The pre-specified McNemar contrasts give T2 (prompt-only vs.\ enforced) $p_{\text{raw}} = 0.45$ under sensitivity (vs.\ $0.45$ original) and T3 (shared-history vs.\ enforced) $p_{\text{raw}} = 0.05$ under sensitivity (vs.\ $0.77$ original). Under Holm-Bonferroni over the three planned tests, neither contrast reaches significance ($p_{\text{adj}} > 0.10$). The qualitative finding that prompt-only and enforced pass rates are statistically indistinguishable holds under both treatments. The directional interpretation of T3, however, flips under sensitivity. With the missing cells treated as failures, enforced-shared-history sits below enforced rather than above. We therefore report T3 as inconclusive rather than as evidence for the shared-history condition. The $3.6\times$ reduction in \emph{verifier-modifies-code} events under enforcement is computed on per-turn events rather than per-run outcomes and is unchanged.

\subsection{Enhanced Solo}
\label{app:enhanced-solo}

\begin{table}[h]
\centering
\caption{Enhanced solo baselines on TeamBench-Mini (28 tasks, gemini-3-flash-preview). Structured prompting does not close the team gap. Solo-CoT is worse than the standard solo.}
\label{tab:enhanced-solo}
\small
\begin{tabular}{lcc}
\toprule
Condition & Avg.\ Partial Score & vs.\ Solo \\
\midrule
Solo (standard) & 0.704 & baseline \\
Solo-CoT        & 0.640 & $-6.4$ \\
Solo-2Pass      & 0.698 & $-0.6$ \\
Full Team         & \textbf{0.754} & $+5.0$ \\
\bottomrule
\end{tabular}
\end{table}

The Solo-CoT condition instructs the single agent to first read the full specification carefully, create a detailed plan, and then implement the solution. Solo-2Pass runs the agent in two sequential phases (specification comprehension, then implementation). Neither condition improves over the minimal solo prompt. On TeamBench-Mini, these two structured-prompting baselines do not close the gap to the full team, which suggests that the measured benefit is not merely a consequence of asking a single model to plan more explicitly.

% \paragraph{Budget-matched Solo runner (released for future studies).} The release includes a budget-matched Solo runner (\texttt{ORACLE\_BUDGET\_MATCHED} in \texttt{harness/ablation.py}, with the convenience wrapper \texttt{scripts/run\_budget\_matched\_solo.py}) that gives the single agent the team's full combined turn budget ($15+25+10=50$ turns) on the $25$-task cross-provider subset. We do not include this condition in the present analysis. We document it here so that follow-up work can use the same runner to test whether the team gain is preserved at matched compute.

\section{Implementation Details and Examples}
\label{app:impl-details}

\subsection{Agent tools}
\label{app:tools}

Table~\ref{tab:tools} lists the four tools available in TeamBench and their access permissions per role. All tools follow a unified interface. Each accepts typed parameters and returns structured output (stdout, stderr, exit code). Tool access is enforced programmatically. Attempts to use a disallowed tool return a permission denied error.

\begin{table}[h]
\centering
\caption{Tool definitions and per-role access permissions. \cmark\,= allowed, \xmark\,= denied. The Planner and Verifier cannot execute arbitrary commands or modify the workspace.}
\label{tab:tools}
\small
\resizebox{\columnwidth}{!}{%
\begin{tabular}{lp{5.2cm}ccc}
\toprule
Tool & Description & Planner & Executor & Verifier \\
\midrule
\texttt{read(path)} & Read file contents from allowed directories & \cmark & \cmark & \cmark \\
\texttt{write(path, content)} & Write to a file in allowed directories & \xmark & \cmark & Attest.\ only \\
\texttt{run(cmd)} & Execute a shell command in the workspace & \xmark & \cmark & \xmark \\
\texttt{send\_message(to, content)} & Send a message to another role & \cmark & \cmark & \cmark \\
\bottomrule
\end{tabular}%
}
\end{table}

\paragraph{File access boundaries.}
Each role has a distinct set of allowed read and write roots, enforced via absolute path prefix checks. The boundaries here match the matrix in Appendix~\ref{app:role-permissions}.
\begin{itemize}[nosep]
    \item \textbf{Planner.} Reads the full specification (\texttt{spec.md}) and the task directory. Reads and writes \texttt{messages/}. No read or write access to the workspace or to \texttt{reports/}.
    \item \textbf{Executor.} Reads \texttt{brief.md}, the workspace, \texttt{messages/}, and writes to the workspace, \texttt{messages/}, and \texttt{reports/} (its own test logs). No read access to the full specification.
    \item \textbf{Verifier.} Reads the specification, the workspace (read-only), and \texttt{reports/} (read-only, the Executor's test logs). Writes \texttt{messages/} and \texttt{attestation.json}. No write access to the workspace, or \texttt{reports/}, and no command execution. The Verifier therefore cannot run tests itself and must inspect the Executor's test logs to require evidence.
\end{itemize}

When running in Docker mode, these boundaries are enforced via container bind mounts. Each role's container only mounts the directories it is permitted to access. Shell escape attempts cannot circumvent the restrictions because the disallowed paths are not present in the container filesystem.

\subsection{Core system prompts}
\label{app:prompts}

\promptbox{Planner System Prompt}{You are the Planner. You have access to the full task specification. Your job is to understand the requirements, decompose the goal, and create a clear plan. You CANNOT execute commands or modify the workspace. You MUST communicate your plan to the Executor by calling the \texttt{send\_message} tool. Highlight hidden constraints and edge cases the Executor might miss.}

\promptbox{Executor System Prompt}{You are the Executor. You can run commands and edit files in the workspace. You only have access to a brief summary of the task. Follow the Planner's instructions carefully. For file reads and writes, use paths relative to the workspace (for example \texttt{app/main.py}). When done with your work, send a message to the Verifier and output DONE. Ask the Planner for clarification if requirements are unclear.}

\promptbox{Verifier System Prompt}{You are the Verifier. You independently verify whether the task was completed correctly. You have read-only access to the workspace and reports. You have access to the full task specification for checking compliance. You CANNOT execute commands or modify the workspace. Your job is to check every requirement, identify violations, and produce \texttt{attestation.json}. If requirements are not met, send feedback to the Executor and set \texttt{verdict='fail'}. Only set \texttt{verdict='pass'} when ALL requirements are satisfied. When done, output DONE.}

\promptbox{Solo System Prompt}{You are a software engineer. You have access to the full task specification and can execute any command. Complete the task to the best of your ability.}

\noindent The solo has access to all four tools with no restrictions on file paths or command execution. Prompts are intentionally concise. Structural enforcement of tool access (Appendix~\ref{app:tools}) carries most of the role-separation burden rather than relying on prompt compliance.

\subsection{Task examples: when teams help and when they hurt}
\label{app:example}

Two contrasting tasks illustrate the conditional value of agent teams measured throughout the paper.

\paragraph{When the team is necessary (HIGH-TNI): MULTI1\_polyglot\_build.}
The workspace contains a multi-language project (Python orchestration, a Go binary, a Node CLI) with a build that fails because the language toolchains expect mutually inconsistent file layouts. Only the specification (visible to the Planner) names which layout is canonical and which two are vestigial. A Restricted agent without specification access reads three plausible file trees and picks the wrong one, scoring $0.10$. The full-access Solo agent does see the spec but spends most of its turn budget on the wrong sub-build before pivoting, scoring $0.45$. The full team passes ($1.00$): the Planner extracts the canonical-layout instruction, the Executor follows it directly, and the Verifier rejects two intermediate attempts that broke the cross-language build. $\tni = 2.00$ on this task: the team \emph{exceeds} what the single agent recovers from full access.

\paragraph{When the team hurts (TEAM-HURTS): GH6\_queryset\_union.}
A real Django bug report on QuerySet union handling. The Solo agent scores $0.25$ partial by returning a working but shallow fix, and the Restricted agent matches at $0.25$. The Team-No-Eval condition reaches $0.33$. The full team scores $0.25$, lower than No-Eval. Trace inspection shows the Verifier reading an Executor patch that would have improved the grader score, asking for an overly defensive guard, and the Executor over-correcting in a way that breaks one of the original tests. The task is small enough that the Executor would have submitted the right answer, but the Verifier's intervention introduces rework that displaces correct work. The pattern, role-induced rework on tasks the single agent partially solved, is exactly the capability-floor inversion measured at the aggregate level in \S\ref{sec:capability}: structure helps when the Executor lacks any starting point and hurts when it would have succeeded alone.

% \clearpage
% \newpage
% \input{checklist.tex}

\end{document}